%% file: main.tex
\documentclass[11pt]{article}

\usepackage[preprint]{acl}

\usepackage{times}
\usepackage{latexsym}
\usepackage[T1]{fontenc}
\usepackage[utf8]{inputenc}
\usepackage{microtype}
\usepackage{inconsolata}
\usepackage{graphicx}
\usepackage{booktabs}
\usepackage{array}
\usepackage{multirow}
\usepackage{amsmath}
\usepackage{amsfonts}
\usepackage{nicefrac}
\usepackage{xcolor}
\usepackage{colortbl}
\usepackage{subcaption}
\usepackage[most]{tcolorbox}
\usepackage{tikz}
\usetikzlibrary{positioning,arrows.meta}

\input{macros}

\title{Evaluating Communicative Success in Machine-Translated Conversation}

\author{%
  Faiz Ghifari Haznitrama \quad Alice Oh \\
  School of Computing, KAIST \\
  \texttt{haznitrama@kaist.ac.kr}
}

\begin{document}
\maketitle

\input{sections/00_abstract}
\input{sections/01_introduction}
\input{sections/02_related_work}
\input{sections/03_method}
\input{sections/04_results}
\input{sections/04c_multiturn}
\input{sections/05_analysis}

\input{sections/06_judge_validation}

\input{sections/07b_conclusion}
\clearpage
\input{sections/09_limitations}

\input{sections/10_ethics}

\bibliography{custom}

\clearpage
\appendix

\input{appendix/A_terminology}

\input{appendix/B_dataset_construction}
\input{appendix/C_models_prompts_scoring}

\input{appendix/D_judge_validation_detail}
\input{appendix/E_human_calibration_detail}
\input{appendix/F_language_direction_results}

\input{appendix/G_multiturn_study}
\input{appendix/H_prompt_ablation}
\input{appendix/I_full_prompts}
\input{appendix/J_qualitative_examples}

\end{document}

%% file: macros.tex
\newif\ifdraftnotes
\draftnotesfalse

\newif\ifshowrev
\showrevfalse

\usepackage[normalem]{ulem}   %
\newif\ifshowrv
\showrvfalse
\definecolor{rvred}{RGB}{220,0,0}

\definecolor{weakcell}{HTML}{F1D9A8}
\definecolor{badcell}{HTML}{B84A3A}
\newcommand{\weak}[1]{\cellcolor{weakcell}#1}
\newcommand{\bad}[1]{\cellcolor{badcell}\textcolor{white}{\textbf{#1}}}

%% file: sections/00_abstract.tex
\begin{abstract}
Interpreter agents built on machine translation (MT) increasingly mediate live conversation, yet existing MT metrics cannot measure whether the translated message is understood and enables communication.
We introduce a reusable, context-grounded checklist framework that evaluates interpreter-mediated text conversation across semantic, pragmatic, and cultural-social layers, capturing aspects missed by conventional MT metrics to serve as a closer proxy to evaluate communicative success. 
The framework supports both single-turn evaluation and interactive multi-turn simulations, scoring individual turns alongside overall conversational outcomes.
We validate our approach via human validation, controlled perturbations, and cross-judge comparisons.
Using this framework, we benchmark 10 interpreter setups across Arabic, Bengali, Indonesian, and Korean, covering 6,000 OpenSubtitles scenarios in 12 single-turn directions and multi-turn dialogues across all 6 language pairs.
Under the main judge and two independent checklist generators, pragmatic and cultural-social criteria are harder than semantic, while conventional metrics overlook critical failures among stronger interpreters.
Finally, prompt ablations show that scenario context and cultural adaptation drive LLM advantages over dedicated MT models.
Together, our framework and benchmark offer a more faithful, communication-grounded proxy for evaluating conversational interpreter agents.
\end{abstract}

%% file: sections/01_introduction.tex
\section{Introduction}
\label{sec:introduction}

Two people who share no language can now hold a conversation through machine translation (MT), typing in a chat app or speaking to a phone that interprets as they talk.
That role has long depended on human interpreters, who understand conversational context beyond words and deliver the intended communicative purpose while adapting the translation for the target user.
This aligns with Skopos translation theory \citep{vermeer-1989-skopos}, where the purpose of a translation drives the translation process, and LLMs and dedicated MT models are now being tried in this interpreter role \citep{sperber-2020-speechtranslation,robinson-2023-chatgptmt}.

\begin{figure}[t]
  \centering
  \includegraphics[width=\columnwidth]{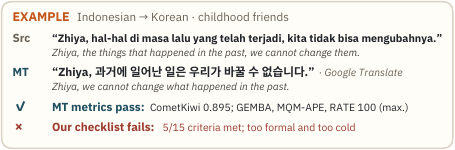}
  \caption{Sample (ind~$\to$~kor) that MT metrics score at or near their maximum and our checklist fails.}
  \label{fig:example-translation}
\end{figure}

However, in complex settings like conversation, where implicit aspects such as intent, tone, naturalness, and cultural appropriateness matter most, these MT models often fail.
Existing MT metrics such as BLEU \citep{papineni-2002-bleu}, chrF \citep{popovic-2015-chrf}, model-based metrics like BERTScore \citep{zhang-2020-bertscore} and COMET/CometKiwi \citep{rei-2020-comet,rei-2022-cometkiwi}, and MQM-based metrics \citep{lommel-mqm,freitag-2021-mqm,kocmi-2023-gemba,kocmi-2023-gembamqm,lu-2024-mqmape} are not built to capture these implicit aspects in fine-grained detail.
Figure~\ref{fig:example-translation} shows a real case from our data, where one of the most popular translation applications fails to deliver a proper translation while scoring highly on CometKiwi.
In conversation, translation accuracy and communicative success are related but distinct, and the gap between them is where a translation model is most likely to fail \citep{moghe-2023-extrinsic}. This is more pronounced for lower-resource languages and for pairs without English, where direct translation remains underexplored.

This paper studies MT models in that interpreter role and builds an evaluation framework around the resulting gap.
Each translated turn is evaluated against a scenario-specific 3-layer checklist covering semantic, pragmatic function, and cultural-social constraints, decomposing communicative success into small, checkable claims \citep{ribeiro-2020-checklist}.
We apply the framework to 10 interpreter setups across 4 languages (Arabic, Bengali, Indonesian, Korean), 6 pairs, and 12 directions over 6{,}000 text-based scenarios built from OpenSubtitles movie and TV dialogue \citep{lison-2016-opensubtitles}, and ground the checklist generator and the generated scenario context in a one-time, pair-specific cultural context of verifiable linguistic and cultural differences between the two languages.
We make four contributions:
\begin{enumerate}
  \item A 3-layer checklist framework for single- and multi-turn translated conversation. For LLM interpreters it is close to orthogonal to existing MT metrics (Spearman $\rho\leq0.31$) and catches failures they miss (\S\ref{sec:analysis-mtmetrics}).
  \item A reusable data-to-evaluation pipeline and benchmark: 6,000 scenarios, 60,000 evaluated translations, and multi-turn dialogue in all 6 pairs.
  \item Validation of every LLM-generated component: simulated users, checklists (human ratings and an independent generator), and the judge (human and adversarial tests, \S\ref{sec:judge-validation}).
  \item Evidence that pragmatic and cultural-social criteria are harder to preserve than semantic ones for every setup, and that scenario context and cultural adaptation drive most of the LLM advantage over dedicated MT (\S\ref{sec:analysis-prompting}).
\end{enumerate}

%% file: sections/02_related_work.tex
\section{Related Work}
\label{sec:related-work}

\paragraph{MT Metrics.}
Reference-based metrics such as BLEU \citep{papineni-2002-bleu}, chrF
\citep{popovic-2015-chrf}, and BERTScore \citep{zhang-2020-bertscore} measure translation
similarity, while learned quality-estimation metrics such as COMET and CometKiwi
\citep{rei-2020-comet,rei-2022-cometkiwi} predict human judgments without a reference. MQM
structures human error annotation \citep{lommel-mqm} at scale \citep{freitag-2021-mqm}, and LLM
evaluators such as GEMBA, GEMBA-MQM, MQM-APE, and RATE extend these with generated judgments or
rubrics \citep{kocmi-2023-gemba,kocmi-2023-gembamqm,lu-2024-mqmape,tian-2026-rate}. All still assess
translation quality, leaving communicative success unmeasured. 
That distinction motivates extrinsic MT evaluation \citep{moghe-2023-extrinsic} and matters
most in speech and real-time settings, where an interpreter agent has to keep a conversation moving \citep{sperber-2020-speechtranslation}. 
We use a granular checklist as a proxy for communicative success.

\paragraph{LLM-as-Judge Reliability and Checklist-Based Evaluation.}
Using an LLM to grade generated text is common \citep{zheng-2023-mtbench,liu-2023-geval}, but
documented failure modes include position and self-preference bias
\citep{wang-2023-notfair,panickssery-2024-selfpreference}, preference for longer or more
confidently styled responses \citep{wu-2023-styleoversubstance}, and acquiescence to surface
wording \citep{hwang-2026-framingbias}. Mitigations include panels of diverse judges
\citep{verga-2024-poll} and decomposing one scalar judgment into small, independently
checkable claims. That decomposition has precedents in behavioral NLP testing
\citep{ribeiro-2020-checklist} and factuality evaluation
\citep{min-2023-factscore,wei-2024-safe}, and was recently shown to be among the most reliable ways
to use an LLM as a judge \citep{cho2026askdontjudgebinary}. 
We extend these ideas to translation evaluation while extensively validating the judge and other LLM-generated components.

\paragraph{Low-Resource, Culturally-Grounded, and Non-Literal MT.}
Resource inequality shapes NLP research
\citep{joshi-2020-stateandfate,ranathunga-2023-survey}, motivating participatory MT
\citep{nekoto-2020-masakhane} and multilingual benchmarks
\citep{goyal-2022-flores101,ahuja-2023-mega}, while culturally grounded NLP shows how systems
encode one viewpoint as a default \citep{hershcovich-2022-crosscultural,naous-2024-culturalbias}.
Work on politeness and register control
\citep{sennrich-2016-politeness,rao-2018-gyafc,madaan-2020-politeness,hwang-2021-honorific}
and on idiomatic language \citep{dankers-2022-idiom,baziotis-2022-idiomeval,liu-2023-maps,wu2026literaltranslationevaluatingcultural}
shows the difficulty of translating meaning beyond literal content, and Skopos theory gives purpose
a central role in translational action \citep{vermeer-1989-skopos}, which specification-aware
translation makes explicit \citep{kayano-2025-specification}. We connect these by
measuring semantic, pragmatic, and cultural-social success from conversational context.

\begin{figure*}[t]
  \centering
  \includegraphics[width=\textwidth]{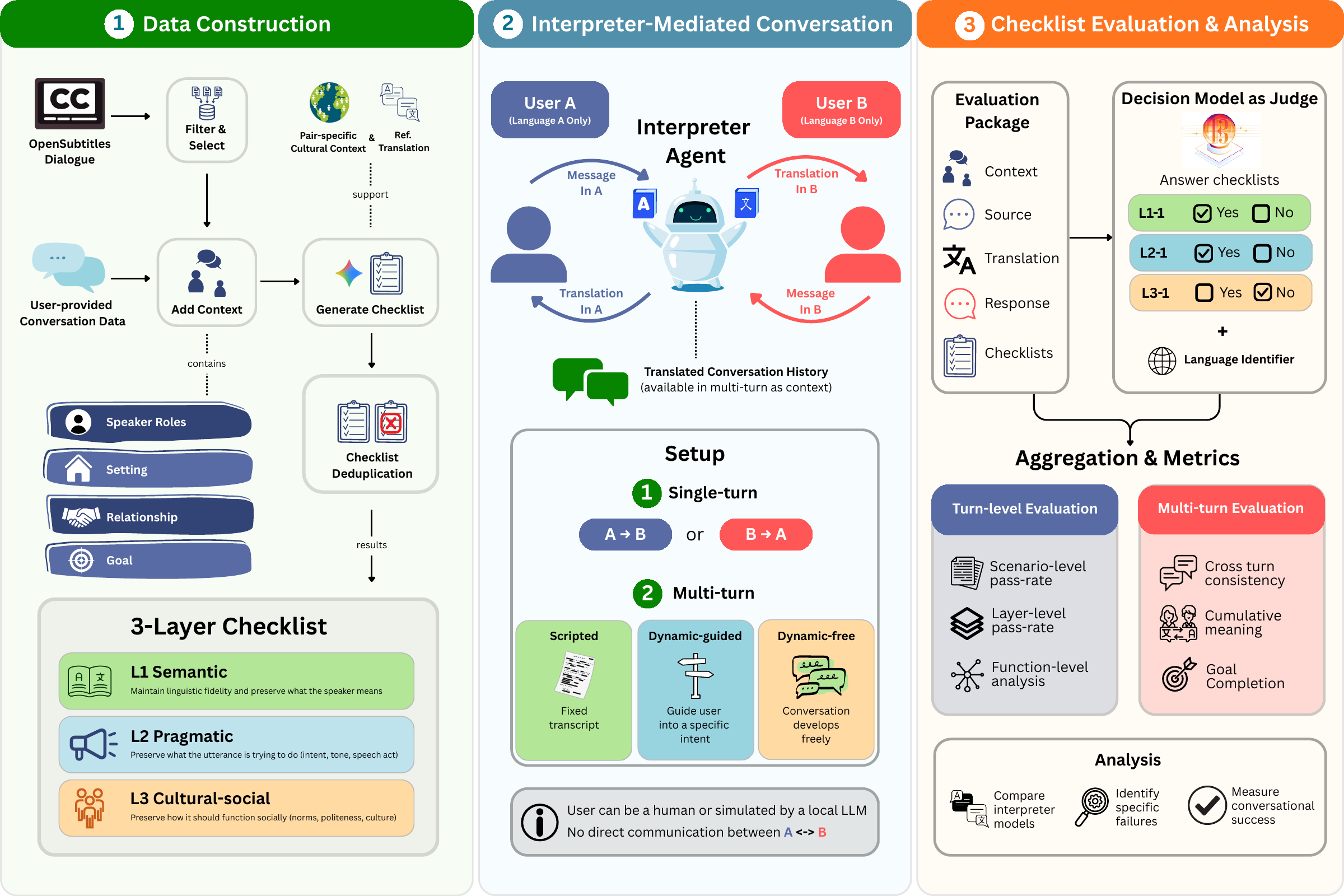}
  \caption{Overview of the conversational evaluation framework, with full data-to-evaluation pipeline.}
  \label{fig:architecture}
\end{figure*}

%% file: sections/03_method.tex
\section{Conversational Evaluation Framework}
\label{sec:method}

\subsection{Data Construction}
\label{sec:data-languages}
\label{sec:data-construction}

Our framework covers a complete data-to-evaluation pipeline for interpreter-mediated conversation
(Figure~\ref{fig:architecture}). We build it from OpenSubtitles \citep{lison-2016-opensubtitles}, a movie- and
television-subtitle corpus under the ODC-By 1.0 license, using 4 languages (Arabic \textsc{arb},
Bengali \textsc{ben}, Indonesian \textsc{ind}, and Korean \textsc{kor}). A bilingual alignment
check and heuristics based on LaBSE similarity \citep{feng-etal-2022-language}, length ratio, and word
coverage yield segments that are cleaner, lower-risk, aligned, and challenging enough to separate
literal fidelity from communicative success. Heuristic features come from each language's
linguistic characteristics and cultural context, plus language-agnostic ones such as script,
length, and pronoun asymmetries.
Manually weighting these, we select the top 500 aligned segments per unordered pair. Bidirectional
augmentation yields 6,000 scenarios. Segments removed by the post-hoc safety filter (Appendix~\ref{sec:appendix-content-filter}) are replaced by the next-ranked ones that pass it.
 We then augment each scenario with
conversation context such as speaker roles, setting, relationship, and communicative purpose, drawn
from up to 15 preceding segments
(Appendix~\ref{sec:appendix-difficulty-scoring}).

\subsection{Communicative-Goal Checklists}
\label{sec:method-checklists}

At the core of the evaluation, we generate checklists as a proxy for communicative success.
A checklist generator first analyzes the setting, speech act, speaker relationship, register, and
pair-specific cultural context, then writes scenario-specific criteria with access to the source text,
conversational context, and reference translation if available. Each
scenario has binary criteria at three scored layers. \textbf{L1 semantic} checks propositions,
entities, quantities, and grammatical distinctions when they alter source meaning.
\textbf{L2 pragmatic} checks whether the intended speech act, tone, and interactional purpose are
preserved naturally. \textbf{L3 cultural-social} checks register, honorifics, face
management, relational stance, and other context-dependent expectations.

Criteria across all three layers are concrete yes/no claims, with yes consistently denoting that
the requirement is met. 
The generator produces 3 sets of criteria for each single-turn scenario, merged with LaBSE
\citep{feng-etal-2022-language} at cosine similarity 0.72 and deduplicated again at 0.80 before scoring, and one set per turn and conversation in multi-turn data.
Across the 6,000 scenarios, L1, L2, and L3 make up 18.1\%, 37.8\%, and 44.0\% of all criteria (Table~\ref{tab:layer-criteria-distribution}).
Criteria testing the same recurring evaluation question (e.g.\ ``uses polite Korean speech level'') are pooled across scenarios into a \emph{function}, the unit used later for function-level analysis and human calibration (Appendix~\ref{sec:appendix-terminology}).

We use Gemini 3.1 Pro \citep{gemini31-modelcard} with high thinking effort as the generator.
The cultural context for each pair was drafted by a locally run LLM agent (Qwen3.6-27B \citep{qwen3.6-27b}) 
that searched published literature on cultural and linguistic gaps between the two languages, 
citing its sources, and reviewed by one native speaker per language (Appendix~\ref{sec:appendix-annotation-protocol}).
Prompts and examples are in Appendix~\ref{sec:appendix-prompts}.

\subsection{Interpreter-Mediated Conversation}
\label{sec:method-architecture}

The framework involves three actors. \textbf{Two users} each produce an utterance in one language
and only observe translated messages from the other, and both can be actual humans or simulated by an
LLM. An \textbf{InterpreterAgent} bridges them by translating and relaying messages according to a
configurable translation brief. A \textbf{Judge} observes the interaction and evaluates both
each turn's translation and the conversation overall, using the conversation context, original
message, and recipient response against the checklists.

\label{sec:method-multiturn}
\textbf{Setup.} In single-turn evaluation, an utterance is translated once and the recipient
replies in the target language using only the translation and its own role context. The multi-turn
setup has 3 modes. 
\textbf{Scripted} replays six-turn OpenSubtitles dialogue, where consecutive lines of one speaker form one turn and each turn's response is fixed.
\textbf{Dynamic-guided} simulates users from a context while following a per-turn intent
outline based on the user's role context. \textbf{Dynamic-free} removes the role context and
lets users respond naturally. 
Multi-turn evaluation retains turn-level L1--L3 criteria and adds
conversation-level criteria for cross-turn consistency, cumulative meaning, and goal completion.

\subsection{Evaluation and Metrics}
\label{sec:method-judge}

The judge receives the context, source text, translated text, recipient response, and L1--L3
checklist, 
and returns the probability $P_c$ that each criterion $c$ is met. 
A criterion counts as met when $P_c\geq\tau$, with threshold $\tau=0.5$ for all languages, layers, and settings (Appendix~\ref{sec:appendix-clef}).
For scenario $s$, we define
\[
  \begin{aligned}
  C_s &= L1_s\cup L2_s\cup L3_s,\\
  p_s &= |C_s|^{-1}\sum_{c \in C_s}\mathbf{1}[P_c\geq\tau].
  \end{aligned}
\]
For the single-turn benchmark, the pass rate is the unweighted mean of $p_s$ over all
6,000 directed scenarios, and layer rates average scenarios within each layer.
Appendix~\ref{sec:appendix-criteria-weighting} provides results for a difficulty-weighted
alternative. Multi-turn evaluation applies the same calculation per turn and scores the
conversation-level checklist once against the transcript.

GlotLID \citep{glotlid-2023} checks the translation language.
For Arabic and Indonesian targets, GlotLID output is checked against a curated
set of accepted dialect/macrolanguage codes (e.g.\ Egyptian and Moroccan for Arabic, Malay and
Betawi for Indonesian) to avoid false negatives. A strict sensitivity analysis instead assigns
zero pass rate when LID rejects a translation (Appendix~\ref{sec:appendix-lid}).

\input{sections/04b_main_results_figure}

\subsection{Experimental Setup}
\label{sec:experimental-setup}
\label{sec:data-models}
\label{sec:data-judge}

\textbf{Interpreters.} The 10 interpreter setups are Gemini 3.1 Pro and Gemini 3.1 Flash Lite
\citep{geminiteam-2023,gemini31-modelcard}, DeepSeek V4 Pro and Flash
\citep{deepseekv4-2026}, GPT-5.4 Mini \citep{gpt54mini-blog}, Qwen3.5 Flash
\citep{qwen3-2025}, Tiny Aya \citep{salamanca2026tinyayabridgingscale}, Google Translate
\citep{johnson-2017-googlenmt}, NLLB-200 3.3B \citep{nllb-2022}, and SeamlessM4T v2 Large
\citep{seamlessm4t-2023}, covering LLM and dedicated MT models and both closed and
open-source setups. By default the interpreter uses a cultural-adaptation brief, which asks for a natural,
culturally adapted translation of the speaker's intent and includes the scenario context
(Appendix~\ref{sec:appendix-prompts}), except for the dedicated MT models, which receive only the source text and language pair. 

\textbf{Simulated Users and Judges.} Simulated users respond with one locally served model per
target language, namely Qwen SEA-LION v4 8B VL \citep{qwen3vl-2025,sealion-v4-modelcard} for Indonesian,
EXAONE 3.5 7.8B Instruct \citep{research-2024-exaone35} for Korean, Command R7B Arabic
\citep{cohere-2025-commandr7barabic} for Arabic, and TigerLLM 9B Instruct
\citep{raihan-2025-tigerllm} for Bengali.
The judge is Clef 27B \citep{clef-2026}, a decision model that returns a probability for each yes/no criterion. It runs zero-shot in 4-bit precision on local GPUs, with one call per scenario covering the whole checklist.
Gemini 3.1 Pro, GPT-5.4, and DeepSeek V4 Pro serve as comparison judges (\S\ref{sec:judge-validation}).
Each setup produced one output per scenario.
Confidence intervals use 10,000 paired bootstrap resamples of scenarios, stratified by direction
(Appendix~\ref{sec:appendix-model-roster}).

\textbf{Data.} The single-turn study uses 6,000 directed scenarios, with checklists averaging 16.1
criteria per scenario (range 7--34) on all 10 setups. 
Scripted multi-turn uses 600 real six-turn OpenSubtitles dialogues, 100 per pair, with
speaker turns aligned across the subtitles of the two languages (Appendix~\ref{sec:appendix-multiturn-pilot}).
It covers
5 interpreter setups (Gemini 3.1 Flash Lite with and without history, Qwen3.5 Flash, NLLB-200,
Tiny Aya) over 3,000 setup-conversations. Dynamic mode covers 2 interpreters (Gemini 3.1 Flash Lite and
Qwen3.5 Flash), guided and free, over 1,000 conversations. Turn-level checklists average
8.5 L1--L3 criteria per scripted turn (range 5--17) and 10.4 per dynamic turn (range 6--22), and conversation-level
checklists 9.2 (range 6--18).

%% file: sections/04b_main_results_figure.tex
\begin{figure*}[t]
  \centering
  \includegraphics[width=\textwidth]{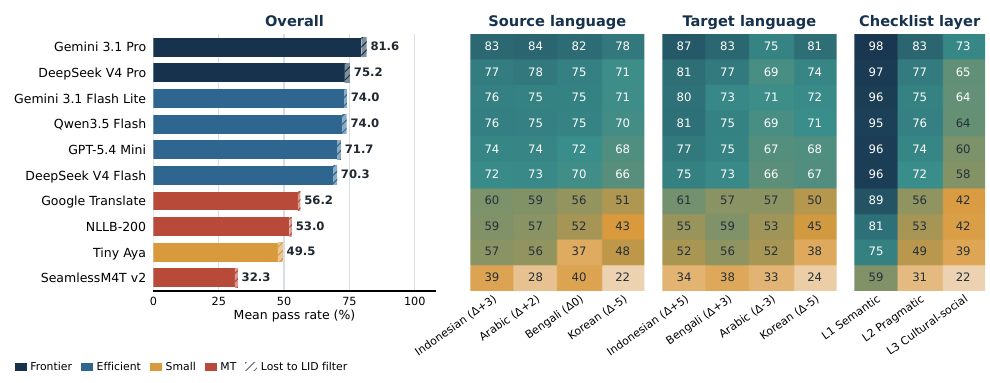}
  \caption{Mean pass rates by interpreter, language (source and target), and checklist layer. $\Delta$ is a language's mean deviation from each interpreter setup's own overall pass rate, averaged across all 10
  setups (\S\ref{sec:results-languages}).}
  \label{fig:central-results}
\end{figure*}

%% file: sections/04_results.tex
\section{Results}
\label{sec:results}

\subsection{Conversation Success Across Setups}
\label{sec:results-headline}

Figure~\ref{fig:central-results} reports the single-turn scenario pass rate for all 10 interpreter
setups by language and checklist layer, from 81.6\% for Gemini 3.1 Pro to 32.3\% for SeamlessM4T v2. Under their default setups, most LLM interpreters,
except the smallest, Tiny Aya, significantly outperform the dedicated MT models. 
Every pair of setups differs significantly except Flash Lite and Qwen3.5 Flash, and efficient LLMs sit within 3.8 points of each other, while frontier models are stronger but considerably more expensive.
However, LLMs use the scenario context and the cultural-adaptation brief, while the dedicated MT
models get only the source text, which we discuss further in \S\ref{sec:analysis-prompting}.

\label{sec:results-languages}%
Language effect is significant in both roles (target $p<0.001$, source $p<0.001$). Indonesian is easiest and Korean hardest in both roles, and Arabic is harder to translate \emph{into} than \emph{from}, which we hypothesize as an adaptation problem (Appendix~\ref{sec:appendix-stats-source-lang}).

\begin{figure*}[t]
  \centering
  \begin{subfigure}[t]{0.49\textwidth}
    \centering
    \includegraphics[width=\linewidth]{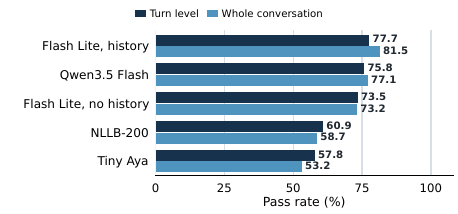}
    \caption{Real scripted dialogue.}
    \label{fig:multiturn-main}
  \end{subfigure}\hfill
  \begin{subfigure}[t]{0.49\textwidth}
    \centering
    \includegraphics[width=\linewidth]{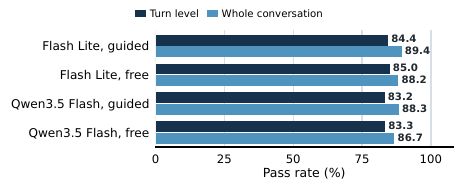}
    \caption{Dynamic conversations by guidance.}
    \label{fig:multiturn-dynamic}
  \end{subfigure}
  \caption{Multi-turn results pooled over pairs. Per-pair results are in Appendix~\ref{sec:appendix-multiturn-pilot}. }
  \label{fig:multiturn-combined}
\end{figure*}

\subsection{Layer and Function Findings}
\label{sec:results-layers}

L2 and L3 are harder than L1 for every setup, which also holds under Gemini 3.1 Pro as judge, with GPT-5.4 checklists, and without the generator's cultural failure-point grounding (Appendices~\ref{sec:appendix-judge-choice} and~\ref{sec:appendix-layer-ablation}). 
Which of L2 and L3 is lower depends on the generator, and after calibration to the target-only 
annotators L1 and L2 are level while L3 stays below L1 for 9 of 10 setups (Appendix~\ref{sec:appendix-cascade}). 
Preserving propositional content does not guarantee that an
interpreted utterance also preserves its conversational function or social fit, and setups
finishing close together overall can still get there through different layer profiles, so we compare
setups by layer. Function pass rates also fall from L1 to L3 in every target language, and individual functions
 expose language-specific weaknesses, such as casual register in Arabic and speech-level formality and emotion-marking endings in Korean (Appendix~\ref{sec:appendix-function-results}).

%% file: sections/04c_multiturn.tex
\section{Interactive Multi-Turn Evaluation}
\label{sec:results-multiturn}

We run the framework over six-turn conversations to evaluate interpreter agents in actual conversation
(\S\ref{sec:method-multiturn}). 
Pooled across pairs (Figure~\ref{fig:multiturn-main}), we observe similar performance compared to 
single-turn results on real scripted dialogue for LLMs ($-0.6$ to $+3.6$ differences),
but improvements for NLLB-200 and Tiny Aya ($+7.9$ and $+8.3$ points). However, conversation-level pass rates
for LLMs are similar to or higher than turn-level pass rates, while they drop slightly for NLLB-200 and Tiny Aya.
This suggests that LLMs maintain conversational function and social appropriateness across turns
and adjust the translation to fit the direction of the conversation, while dedicated MT models are unable to do so because of their limitations.

\paragraph{Conversation memory.}
Utilizing conversational history improves Flash Lite's turn-level pass rate by 4.2 [3.6, 4.9] points 
over 3,600 matched turns, significant in
all 6 pairs ($+3.5$ to $+5.3$ points per pair). History also improves conversation-level pass rates relative to turn-level,
which is another indication that context helps adjust the translation.

\paragraph{Dynamic conversation.}
In dynamic mode, Flash Lite and Qwen3.5 Flash are evaluated on 73 to 89 conversations per
setup--pair, 500 per setup, of which 336 are guided and 164 free. 
Dynamic turns have no reference translation, so we compare dynamic pass rates only within dynamic mode.
Both setups score 4.4 and 4.6 points higher at conversation level than at turn level (Figure~\ref{fig:multiturn-dynamic}). Guided and free conversations differ by at most 1.6 points pooled, and no per-pair difference is significant (Appendix~\ref{sec:appendix-multiturn-dynamic}).

%% file: sections/05_analysis.tex
\section{Analysis}
\label{sec:analysis}

\begin{figure}[t]
  \centering 
  \includegraphics[width=\columnwidth]{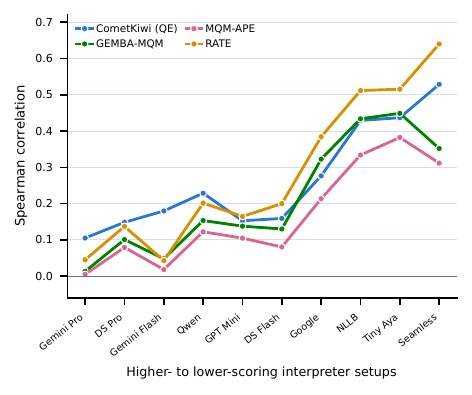}%
  \caption{Within-setup correlation between existing MT metrics and our checklist pass rate.}
  \label{fig:mt-metric-orthogonality}
\end{figure}

\subsection{Relationship to MT Metrics}
\label{sec:analysis-mtmetrics}

If our checklist measured only what standard MT metrics measure, those metrics would closely track its pass rates.
Figure~\ref{fig:mt-metric-orthogonality} shows that within individual setups, the correlation
between our pass rate and CometKiwi, GEMBA-MQM, MQM-APE, and RATE \citep{tian-2026-rate} collapses
as interpreter capability rises (the three LLM-based metrics run on Gemini 3.1 Pro).
These metrics separate translation failure well but miss the finer-grained failures our
checklist is built to catch, and each within-setup correlation is precise to within $\pm0.03$ (Appendix~\ref{sec:appendix-stats-power}). Pooling across setups inflates every metric's 
apparent alignment by rewarding it for common
failure instead of the subtle failures that dominate once outputs are fluent.
Table~\ref{tab:mt-metric-blindspots} makes this concrete for Gemini 3.1 Pro, the strongest interpreter. The translations each metric rates best fail our checklist as often as its translations overall. A top metric score therefore does not make a translation less likely to fail in the conversation.

\textbf{Checklist or context?} Our judge also sees the scenario context, which the MT metrics do not, so the gap in Table~\ref{tab:mt-metric-blindspots} could come from the context alone. We therefore give Clef the same input with one holistic question in place of the checklist (Appendix~\ref{sec:appendix-holistic}). This holistic judge passes 94.4\% of Gemini 3.1 Pro's translations, follows CometKiwi more closely than our checklist does (Spearman $\rho=0.34$ against 0.11), and flags only 10.7\% of the translations in CometKiwi's top quarter that our checklist fails. Decomposing communicative success into criteria is therefore what exposes these failures.

\begin{table}[t]
  \centering
  \caption{Gemini 3.1 Pro translations our checklist fails (pass rate $\leq0.6$), overall and in each metric's top-rated set. Size: the set's share of all translations.}
  \label{tab:mt-metric-blindspots}
  \small
  \begin{tabular*}{\columnwidth}{@{\extracolsep{\fill}}llcc@{}}
    \toprule
    \textbf{Metric} & \textbf{Top-rated set} & \textbf{Size} & \textbf{Fail} \\
    \midrule
    All translations & -- & 100\% & 17.5\% \\
    \midrule
    CometKiwi & top quarter & 25\% & 16.2\% \\
    GEMBA-MQM & perfect score & 57\% & 17.2\% \\
    MQM-APE & perfect score & 67\% & 17.5\% \\
    RATE & perfect score & 79\% & 16.6\% \\
    \bottomrule
  \end{tabular*}
\end{table}

\subsection{Context and Translation Briefs}
\label{sec:analysis-prompting}

We compare six interpreter briefs on 1,108--1,113 paired difficult scenarios for seven LLM setups (Table~\ref{tab:prompt-ablation}).
With a literal brief and no context, the LLMs perform close to dedicated MT (35.9--48.9\% against 41.4--44.0\% for Google Translate and NLLB-200), and removing the literal instruction gains 4.0 points pooled.
Scenario context adds another 4.0 points, after which the neutral brief matches the specification-aware brief, and only the cultural-adaptation brief improves further ($+2.2$ points).
Context and cultural adaptation therefore drive most of the LLM advantage over dedicated MT.

\input{tables/prompt_ablation}

%% file: tables/prompt_ablation.tex
\begin{table}[t]
  \centering
  \caption{Pass rate (\%) on the difficult subset, by interpreter brief and dedicated MT. Columns are target languages. Per-setup breakdown on Appendix~\ref{sec:appendix-prompt-ablation}.}
  \label{tab:prompt-ablation}
  \setlength{\tabcolsep}{3pt}
  \renewcommand{\arraystretch}{1.08}
  \resizebox{\columnwidth}{!}{%
  \begin{tabular}{@{}lccccc@{}}
    \toprule
    \textbf{Brief or setup} & \textbf{All} & \textbf{Ar} & \textbf{Bn} &
    \textbf{Id} & \textbf{Ko} \\
    \midrule
    \multicolumn{6}{@{}l}{\emph{7 LLM interpreters - pooled}} \\
    Literal, no context             & 45.3 & 42.9 & 53.1 & 47.3 & 38.7 \\
    Literal + context               & 47.5 & 44.6 & 55.0 & 50.1 & 41.1 \\
    Neutral, no context         & 49.3 & 44.9 & 54.9 & 53.2 & 44.8 \\
    Neutral + context           & 53.3 & 47.0 & 58.4 & 58.4 & 50.3 \\
    Specification-aware            & 53.3 & 46.8 & 58.7 & 58.4 & 50.1 \\
    Cultural adaptation            & \textbf{55.6} & \textbf{48.0} & \textbf{60.5} & \textbf{61.8} & \textbf{52.8} \\
    \midrule
    \multicolumn{6}{@{}l}{\emph{Dedicated MT (raw text, no context)}} \\
    Google Translate     & 44.0 & 42.7 & 46.5 & 47.7 & 39.6 \\
    NLLB-200 3.3B        & 41.4 & 39.3 & 48.9 & 42.6 & 35.7 \\
    SeamlessM4T v2 Large & 25.4 & 27.4 & 30.3 & 24.5 & 19.5 \\
    \bottomrule
  \end{tabular}%
  }
\end{table}

%% file: sections/06_judge_validation.tex
\section{Checklist and Judge Validation}
\label{sec:judge-validation}

\subsection{Human Validation and Calibration}
\label{sec:checklist-validation}
\label{sec:judge-validation-uncertainty}
\label{sec:judge-validation-human}

We conduct human validation of the checklist and judge with native speakers of the four target languages (Appendix~\ref{sec:appendix-human-annotation}). We first check whether the checklist criteria are meaningful to evaluate translation success. Annotators rate each function's \emph{Meaningfulness} (1--5) with no translations shown. We find that mean \emph{Meaningfulness} exceeds 3 for every target language (3.1--4.27, Appendix~\ref{sec:appendix-annotation-protocol}), which indicates that the criteria are valid to check. We also find that filtering out criteria rated lower than 4 do not change any of the main conclusions in our results.

\input{tables/judge_reliability}

Annotators then gave 3,477 labels on 2,089 criteria in real translations, 800 of them labeled by two or three annotators, and we compare four candidate judges (Clef 27B, Gemini 3.1 Pro, GPT-5.4 \citep{gpt54-systemcard}, and DeepSeek V4 Pro \citep{deepseekv4-2026}) against these labels.
Annotators agree with each other on 72.5\% of items (pairwise Cohen's $\kappa=0.21$, Krippendorff's $\alpha=0.25$), so we count a judge as reliable when its $\kappa$ against the annotators exceeds theirs with each other and adding it as a rater does not lower $\alpha$.
All four judges meet this bar (Table~\ref{tab:judge-reliability}), with $\kappa$ of 0.33--0.40 against the annotators, and adding any one of them as a rater raises $\alpha$ above 0.25.

Propagating the human labels to the full dataset (Table~\ref{tab:cascade-shift} in Appendix~\ref{sec:appendix-cascade}), Clef scores 13.9 to 24.7 points below the target-only annotators in all four languages, with 81--93\% of function--verdict cells calibrated per language. Calibration lowers L1 by 3.9--8.1 points and raises L2 and L3 by 10.2--29.3, narrowing the layer gap.
We expect this moderate $\kappa$ and stricter judge, because annotators see only the target text and checklist, while 18--43\% of functions need the source (Appendix~\ref{sec:appendix-checklist-meaningfulness}).

\subsection{Adversarial Judge Tests}
\label{sec:judge-validation-suite}

We further ran four tests on the four judge candidates to isolate different failure types (Table~\ref{tab:judge-validation-compact-candidate}). \textbf{Translation dropout} measures how much the judge's yes-rate drops when it evaluates without the translation. \textbf{Meaning-preserving
rewording} measures decision instability when only checklist wording changes. \textbf{Polarity
inversion} reverses the expected answer, testing acquiescence to surface
form. \textbf{Planted violations} alter one translation requirement, measuring whether the judge detects localized failures without changing other verdicts.

\input{tables/judge-validation-compact-candidate}

From the tests, we found that LLM judges (Gemini 3.1 Pro most of all) still pass most criteria when they cannot see the translation, means their verdicts rely partly on the scenario and the checklist. Clef bases its verdicts on the translation itself, shown from higher dropout.
DeepSeek V4 Pro tends to answer yes whatever the question asks, which rules it out as a judge, and GPT-5.4 shows the same tendency to a lesser degree.
The LLM judges catch more planted errors and change fewer verdicts when the checklist is reworded. Clef changes fewer unrelated verdicts when an error is planted, and gives the same verdicts across runs, while other judges change 2.2--8.4\% of their verdicts between two runs (Appendix~\ref{sec:appendix-judge-validation-detail}).
We choose Clef as the judge for its independence from the checklist generator and the interpreters, its reproducibility, and its cost, while the main conclusions still hold under Gemini 3.1 Pro (Appendix~\ref{sec:appendix-judge-choice}).

\makeatletter\global\@topnum=0\makeatother
\begin{figure}[t]
  \centering
  \includegraphics[width=\columnwidth]{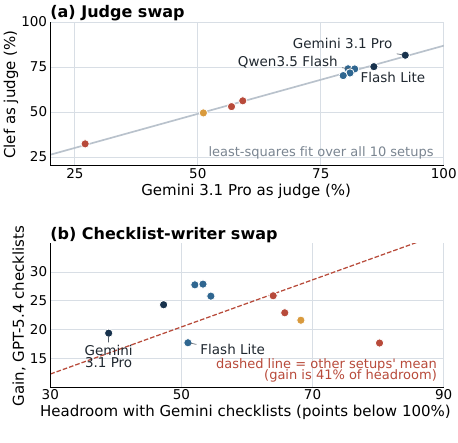}
  \caption{Self-preference validation, one point per interpreter setup (pass rates in points).}
  \label{fig:selfpref-main}
\end{figure}

\subsection{Self-Preference Validation}
\label{sec:judge-validation-shared}

\textbf{Judge.} Clef is post-trained from Qwen3.8-27B \citep{qwen3.8-2026} and Qwen3.5 Flash is one of the interpreter setups, while Gemini 3.1 Pro as secondary judge scores its own translations. Neither judge prefers its own family off the shared trend (Figure~\ref{fig:selfpref-main}a). On identical translations, Clef's gap to the three LLM judges differs by only 0.7 points between Qwen3.5 Flash and Gemini 3.1 Flash Lite (95\% CI $[-0.5, +1.9]$), so any family preference is at most 1.9 points.

\textbf{Checklist generator.} We try using GPT-5.4 to rewrite the checklists of 463 difficult scenarios with the same prompt. Its checklists raise every setup by 18 to 28 points and keep Gemini 3.1 Pro first, with Kendall $\tau=0.867$ against the original ranking under Clef and 0.911 under Gemini 3.1 Pro. Gemini 3.1 Pro closes more of its remaining headroom than the other setups on average (Figure~\ref{fig:selfpref-main}b), which indicates that Gemini-written checklists show no preference for Gemini translations. Flash Lite closes the least among the LLMs because GPT-5.4 criteria penalize its bracketed clarifications (Appendix~\ref{sec:appendix-generator-layers}).

%% file: tables/judge_reliability.tex
\begin{table}[t]
  \centering
  \caption{Agreement with human annotators, with annotators compared to each other in the first row. $\kappa$ is Cohen's kappa, and Krippendorff's $\alpha$ adds the judge as a rater. Unanimous uses items all annotators agree on.}
  \label{tab:judge-reliability}
  \small
  \setlength{\tabcolsep}{2pt}
  \begin{tabular*}{\columnwidth}{@{\extracolsep{\fill}}lcccc@{}}
    \toprule
    \textbf{Rater} & \textbf{Agree (\%)} & \textbf{$\kappa$} & \textbf{$\alpha$} & \textbf{Unanimous (\%)} \\
    \midrule
    Annotators      & 72.5 & 0.21 & 0.25 & -- \\
    Clef 27B        & 72.1 & 0.33 & 0.31 & 84.0 \\
    Gemini 3.1 Pro  & 75.4 & 0.40 & 0.35 & 89.5 \\
    GPT-5.4         & 69.3 & 0.33 & 0.29 & 79.0 \\
    DeepSeek V4 Pro & 77.2 & 0.38 & 0.32 & 89.4 \\
    \bottomrule
  \end{tabular*}
\end{table}

%% file: tables/judge-validation-compact-candidate.tex
\begin{table}[t]
  \centering
  \caption{Adversarial judge tests. Shaded cells mark the weakest aspects of the usable judges, and the red cell marks the weakness that rules a judge out.}
  \label{tab:judge-validation-compact-candidate}
  \setlength{\tabcolsep}{4pt}
  \small
  \begin{tabular}{@{}lcccc@{}}
    \toprule
    \textbf{Judge} & \textbf{\shortstack{Dropout\\$\uparrow$}} & \textbf{\shortstack{Reword\\flip $\downarrow$}}
      & \textbf{\shortstack{Polarity\\drop $\downarrow$}} & \textbf{\shortstack{Planted\\sens.\ $\uparrow$}} \\
    \midrule
    Clef 27B & 26.3 & \weak{7.1\%} & 6.7 & \weak{0.72} \\
    Gemini 3.1 Pro  & \weak{4.2}  & 2.2\% & 5.5  & 0.82 \\
    GPT-5.4         & 14.5 & 2.7\% & \weak{22.8} & 0.90 \\
    DeepSeek V4 Pro & 12.5 & 2.7\% & \bad{60.1} & 0.81 \\
    \bottomrule
  \end{tabular}
\end{table}

%% file: sections/07b_conclusion.tex
\section{Conclusion}
\label{sec:conclusion}

We introduce a 3-layer checklist framework that evaluates interpreter-mediated conversation across semantic, pragmatic, and cultural-social criteria, and apply it to 10 interpreter setups over 6,000 single-turn scenarios and multi-turn dialogue in 4 languages.
Pragmatic and cultural-social criteria are harder than semantic ones for every setup.
The checklist catches failures that existing MT metrics rate highly, especially for strong LLM interpreters, and a judge given the same context without the checklist misses most of them, so the decomposition into criteria is what exposes these failures.
In multi-turn dialogue, conversation history raises Flash Lite's pass rate, and the LLMs keep their pass rate over whole conversations while NLLB-200 and Tiny Aya drop.
Scenario context and cultural adaptation drive most of the LLM advantage over dedicated MT.
The Clef judge agrees with native-speaker annotators at least as well as they agree with each other and shows no self-preference, which keeps the benchmark reproducible at no API cost.
Checklists grounded in the conversation therefore track communicative success more closely than existing MT metrics, especially for strong interpreters those metrics rate as near-perfect.

%% file: sections/09_limitations.tex
\section*{Limitations}
\label{sec:limitations}

\textbf{Human validation.} Our main limitation is that the annotators are not bilingual. We could not find annotators fluent in both languages of a pair, so each annotator judges the translation against the checklist in the target language only, without the source. This lowers human--human and human--judge agreement, and annotators who cannot see the source credit criteria more often than the judge does. Calibrating Clef with their agreement to the full dataset raises Clef's pass rates by up to 24.7 points (Appendix~\ref{sec:appendix-cascade}). An evaluation by bilingual annotators who see the source would separate judge strictness from this missing information, and we leave it to future work.

\textbf{Data contamination.} OpenSubtitles is a common component of LLM pretraining corpora, so every model, including the judge, may have seen this exact corpus during pretraining. A model that memorized subtitle translation patterns could therefore score well without reasoning about pragmatics or culture. Since our approach does not
rely on a single reference translation and decomposes the evaluation into criteria, we expect a limited contamination effect. 

\textbf{Scripted OpenSubtitles dialogue.} Scripted OpenSubtitles dialogue is another limitation. Despite our alignment checks and filtering
heuristics, some segments remain imperfectly aligned, and since most films originate outside these
4 languages, many subtitles are themselves third-language translations, making dialogue less
natural than spontaneous speech. We take this as the best available data for this work, but do not
claim our results generalize beyond scripted dialogue to natural conversation. The framework
scores the text of each turn, so spoken interpreting requires an ASR step, and we
have not tested it on live speech.

\textbf{Simulated users.} Simulated-user replies come from 7--9B monolingual models
served locally, one per target language, and the judge uses these replies as evidence for
some checklists related to pragmatic and cultural-social success. 
We validate their language fidelity and context adherence
(Appendix~\ref{sec:appendix-simuser-validation}). Some of their turns break character with a safety refusal, and excluding these turns barely changes pass rates. We still have not compared their replies with those of real native-speaker recipients, 
so the pragmatic and cultural-social verdicts depend on how a small model reacts to each translation.
Therefore, we do not claim that the simulated users are a perfect proxy for real users. A user study with real cross-lingual participants is the direct test of communicative success and our main next step.

%% file: sections/10_ethics.tex
\section*{Ethical Considerations}
\label{sec:ethics}

All human validation was approved by an institutional review board and conducted with appropriate compensation and informed consent. The released
results of the study contain no personally identifiable information.
The corpus is text only. It carries no audio, video, or likeness data, so no consent for voice or likeness is implicated and the released artifacts pose no deepfake or voice-cloning risk. On provenance, all source material comes from OpenSubtitles \citep{lison-2016-opensubtitles}, redistributed under the ODC-By 1.0 license, and we release only derived scenarios, interpreter setup outputs, and the scripts that produce them.
The OpenSubtitles-derived data may contain offensive content and, although less likely in scripted
movie and television dialogue, personally identifiable information. We therefore apply a rule-based
post-hoc filter to the source and reference data, setup translations, and simulated-user
responses. Because the filter relies on heuristics, some instances may remain undetected in the
released artifacts.
Our results are limited to scripted dialogue and should not be generalized to other domains, particularly real-world use.

We acknowledge the use of AI assistants (Claude Code and Codex) for writing and coding.

%% file: appendix/A_terminology.tex
\section{Terminology}
\label{sec:appendix-terminology}

The paper uses five related but distinct terms for the units the framework operates on.

\begin{itemize}
  \item \textbf{Segment.} One bilingual (source + target) unit selected from OpenSubtitles by the
    difficulty/alignment scoring in \S\ref{sec:data-construction}, giving 500 segments per unordered
    language pair, 3,000 total across the six pairs, with segments removed by the post-hoc safety filter replaced by the next-ranked ones. A segment is direction-agnostic, so it is not
    yet assigned a translation direction.
  \item \textbf{Scenario.} One directed instantiation of a segment (source $\to$ target),
    augmented with conversation context and an L1--L3 checklist. Each segment yields 2 directed
    scenarios (forward and reverse), giving 6,000 scenarios total. This is the $s$ already used in
    \S\ref{sec:method-judge}'s $p_s$/$C_s$ definitions.
  \item \textbf{Turn.} One utterance-exchange within a conversation. The single-turn benchmark
    scores exactly one turn per scenario. Multi-turn evaluation
    (\S\ref{sec:method-multiturn}) scores each turn of a longer conversation plus a separate
    conversation-level checklist.
  \item \textbf{Criterion.} One atomic yes/no claim in a scenario's checklist, the unit for which the judge returns the probability that it is met, thresholded at $\tau$. $C_s$ is the set of criteria scored for scenario $s$.
  \item \textbf{Function.} A cross-scenario \emph{cluster} of criteria that test the same
    recurring evaluation question (e.g.\ ``uses polite Korean speech level''), constructed by the
    embedding-based pooling described in Appendix~\ref{sec:appendix-function-construction}. 
\end{itemize}

%% file: appendix/B_dataset_construction.tex
\section{Dataset Construction, Filtering, and Attrition}
\label{sec:appendix-dataset}
\label{sec:appendix-difficulty-scoring}

\subsection{Worthiness Formula}
\label{sec:appendix-worthiness-formula}

Each candidate OpenSubtitles segment receives a source-side complexity score
$\mathit{src\_complexity}$ (sum of fired per-language feature weights, described below, plus
universal surface signals such as question marks, exclamations, ellipsis, and multi-clause
structure), an alignment-quality score $\mathit{quality}$ (length-ratio and script-ratio checks
against a per-language target, penalized by an alignment-risk heuristic based on
punctuation/digit mismatch and length asymmetry between source and target lines), and a combined
\emph{worthiness} score used to rank and select the hardest 500 segments per unordered pair.
\begin{equation*}
  w = 0.65 \cdot c + 0.35 \cdot q
\end{equation*}
where $w$ is the worthiness score used for ranking, $c$ is the source-side complexity score, and
$q$ is the alignment-quality score. Feature weights are manually set and each is grounded in a
specific citation about why that linguistic phenomenon is translation-critical for its language. Before ranking, a LaBSE similarity filter, calibrated on labeled aligned and misaligned pairs, drops badly misaligned subtitle pairs.

\subsection{Representative Per-Language Features}

Arabic has diglossia and clitic negation \citep{ferguson-1959-diglossia,soudi-2012-challenges},
Korean has honorifics and evidentiality \citep{song-2010-evidentials,hwang-2021-honorific},
Indonesian has discourse particles and voice morphology \citep{mistica-2009-reduplication,wijaya-2022-loh},
and Bengali has echo reduplication and compound verbs \citep{bhattacharja-2010-benglish,mukhopadhayay-2012-compound}.
Table~\ref{tab:difficulty-features} gives one representative fired feature per language with its
weight and grounding citation. The full feature set (roughly 15--20 features per language,
covering honorifics, evidentiality, diglossia, reduplication, compound verbs, register, and
morphological ambiguity) is available in the supplementary scoring code.

\begin{table}[htbp]
  \centering
  \caption{Representative difficulty-scoring features. Korean honorific weighting is grounded in
  \citet{hwang-2021-honorific}, the evidential feature in \citet{chung-2010-evidentials} and
  \citet{song-2010-evidentials}, and Arabic diglossia/negation grounding in
  \citet{ferguson-1959-diglossia} and \citet{soudi-2012-challenges}.}
  \label{tab:difficulty-features}
  \resizebox{\columnwidth}{!}{%
  \begin{tabular}{@{}llc@{}}
    \toprule
    \textbf{Lang.} & \textbf{Feature} & \textbf{Weight} \\
    \midrule
    Korean     & honorific/social marker     & 1.00 \\
    Korean     & retrospective evidential \emph{-deo-} & 0.75 \\
    Indonesian & \emph{meN-} prefix complexity & 0.50 \\
    Arabic     & clitic negation (\emph{m\=a-\ldots-\v{s}}) & 1.00 \\
    Arabic     & broken plural                & 0.75 \\
    Bengali    & honorific marker             & 0.75 \\
    Bengali    & Sadhu (elevated) register    & 0.50 \\
    \bottomrule
  \end{tabular}}
\end{table}

\subsection{Worthiness Score as a Predictor of Pass Rate}
\label{sec:appendix-worthiness-validation}

The feature weights above are grounded in linguistics literature describing what makes a
construction hard for a human, but not in any measurement of what is hard for
an LLM or MT model. Matching the 500 selected segments per pair back to their worthiness score
and correlating against mean pass rate across all 10 interpreter setups, within-pair Spearman correlations
are weak (all six negative and at most 0.16 in magnitude), so the score should
not be read as a validated predictor of setup difficulty.
We take this as an expected consequence of the score's grounding.

\subsection{PII and Offensive-Content Filter}
\label{sec:appendix-content-filter}

A rule-based post-hoc filter scans source and
reference text, translations, and simulated-user responses for high-confidence PII and offensive content, including
ordinary profanity, insults, and slurs in the four study languages and English. Blocking matches
remove both directions of a single-turn base example from every setup. For multi-turn data, a
match removes the entire conversation from every setup. 

The filter removes 188 single-turn
base examples (185 offensive-content, 3 PII) and 40 dynamic
conversations. 
A real scripted dialogue is excluded when any setup's translation of any turn is blocked, and is replaced by an unused dialogue of the same pair and direction. 
For single-turn, we replace each removed base example with the next-ranked segment of the same pair in the same worthiness ordering that passes the filter. 
The replacements are scored with the same checklist generator, interpreters, simulated users, and judges as the rest of the benchmark.
The filter is heuristic and can miss indirect or novel
expressions and overflag context-dependent usage.
Table~\ref{tab:retained-data-by-pair} gives the evaluation units per language pair.

\subsection{Bengali Source Repair}
\label{sec:appendix-bengali-repair}

In the OpenSubtitles release we use, many Bengali lines of the Arabic--Bengali pair lost their dependent vowel
signs, which leaves consonant skeletons that a reader cannot parse. We flag a Bengali line when it has fewer than
0.25 vowel signs per consonant, which flags 410 of the 500 Arabic--Bengali base examples. We replace each with the
next-ranked Arabic--Bengali segment in the same worthiness ordering whose Bengali line passes this check and the
content filter, and score the replacements with the same checklist generator, interpreters, simulated users,
and judges as the rest of the benchmark. The replacements rank lower than the segments they replace (mean worthiness
1.77 against 2.32), and every pair keeps 500 base examples.

\begin{table}[htbp]
  \centering
  \caption{Evaluation units by language pair after content filtering and replacement for single-turn data.}
  \label{tab:retained-data-by-pair}
  \setlength{\tabcolsep}{2.5pt}
  \resizebox{\columnwidth}{!}{%
  \begin{tabular}{@{}lrrrrr@{}}
    \toprule
    \textbf{Pair} & \textbf{\shortstack{Single/\\dir.}} & \textbf{Scripted} & \textbf{\shortstack{Dyn.\\guided}} &
    \textbf{\shortstack{Dyn.\\free}} & \textbf{\shortstack{Dyn.\\total}} \\
    \midrule
    Arabic--Bengali     & 500 & 100 & 56 & 29 & 85 \\
    Arabic--Indonesian  & 500 & 100 & 59 & 29 & 88 \\
    Arabic--Korean      & 500 & 100 & 59 & 30 & 89 \\
    Bengali--Indonesian & 500 & 100 & 50 & 23 & 73 \\
    Bengali--Korean     & 500 & 100 & 52 & 24 & 76 \\
    Indonesian--Korean  & 500 & 100 & 60 & 29 & 89 \\
    \bottomrule
  \end{tabular}}
\end{table}

%% file: appendix/C_models_prompts_scoring.tex
\section{Models and Scoring Protocol}
\label{sec:appendix-model-roster}

\subsection{Interpreter, Generator, and Judge Access}
Table~\ref{tab:model-access} lists the access name and settings of every model. API models otherwise use their provider-default generation settings, and all were accessible during the evaluation and paper submission window.

\begin{table*}[t]
  \centering
  \small
  \caption{Models by access method, with thinking effort for API models and decoding for local ones.}
  \label{tab:model-access}
  \setlength{\tabcolsep}{4pt}
  \begin{tabular*}{\textwidth}{@{\extracolsep{\fill}}llll@{}}
  \toprule
  \textbf{Model} & \textbf{Access name} & \textbf{Effort/decoding} & \textbf{Role} \\
  \midrule
  \multicolumn{4}{@{}l}{\textit{Google AI Studio}} \\
  Gemini 3.1 Pro & \texttt{gemini-3.1-pro-preview} & Low & Interpreter, comparison judge \\
   &  & High & Checklist generator \\
  Gemini 3.1 Flash Lite & \texttt{gemini-3.1-flash-lite-preview} & Minimal & Interpreter \\
   &  & Low & Filtering multi-turn windows \\
  Gemini 3.8 Flash & \texttt{gemini-3.8-flash} & Low & Speaker labels of real dialogue \\
  \midrule
  \multicolumn{4}{@{}l}{\textit{OpenAI API}} \\
  GPT-5.4 & \texttt{gpt-5.4-2026-03-05} & High & Comparison judge \\
   &  &  & Fallback multi-turn generator \\
  GPT-5.4 Mini & \texttt{gpt-5.4-mini-2026-03-17} & Minimal & Interpreter \\
  \midrule
  \multicolumn{4}{@{}l}{\textit{OpenRouter}} \\
  DeepSeek V4 Pro & \texttt{deepseek/deepseek-v4-pro} & Minimal & Interpreter \\
  DeepSeek V4 Flash & \texttt{deepseek/deepseek-v4-flash} & Minimal & Interpreter \\
  Qwen3.5 Flash & \texttt{qwen/qwen3.5-flash-02-23} & Minimal & Interpreter \\
  \midrule
  \multicolumn{4}{@{}l}{\textit{Local, Transformers fp16}} \\
  Tiny Aya & \texttt{CohereLabs/tiny-aya-global} & Greedy, 256 tokens & Interpreter \\
  NLLB-200 3.3B & \texttt{facebook/nllb-200-3.3B} & Beam 4, 512 tokens & Interpreter \\
  SeamlessM4T v2 Large & \texttt{facebook/seamless-m4t-v2-large} & Beam 4, 512 tokens & Interpreter \\
  \midrule
  \multicolumn{4}{@{}l}{\textit{Local, 4-bit (bitsandbytes), four RTX 2080 Ti 11\,GB}} \\
  Clef 27B \citep{clef-2026} & \texttt{Cloudflare/clef} & Zero-shot & Judge, one call per scenario \\
  \midrule
  \multicolumn{4}{@{}l}{\textit{Google Cloud API}} \\
  Google Translate & -- & -- & Interpreter \\
  \midrule
  \multicolumn{4}{@{}l}{\textit{Simulated users, local, llama.cpp, 4-bit (\texttt{Q4\_K\_M}), one RTX 4060 Ti 16\,GB}} \\
  Qwen SEA-LION v4 8B VL & \texttt{qwen-sea-lion-v4-8b-vl} & Default, 32K context & Indonesian user \\
  EXAONE 3.5 7.8B & \texttt{exaone-3.5-7.8b-instruct} & Default, 16K context & Korean user \\
  Command R7B Arabic & \texttt{c4ai-command-r7b-arabic-02-2025} & Default, 16K context & Arabic user \\
  TigerLLM 9B & \texttt{tigerllm-9b-it} & Default, 16K context & Bengali user \\
  \bottomrule
  \end{tabular*}
\end{table*}

\subsection{Clef Judge}
\label{sec:appendix-clef}
Clef receives the Judge Prompt of Appendix~\ref{sec:appendix-prompts} with three parts removed: the
verification checklist block, the request for a reasoning, and the JSON output instructions. Each checklist
criterion becomes one boolean field of Clef's output schema, and Clef returns $P_c$ for each field in one
call per scenario. The threshold is $\tau=0.5$, the standard decision threshold for a probability,
with no tuning on our data.
For 26 records whose inputs exceeded the judge's GPU memory, we cut the simulated user's response to its first 2,000 characters, and the translation is kept in full. Three Qwen3.5 Flash outputs that returned their reasoning in place of a translation are cut to their first 300 characters.

\subsection{Simulated Users}
\label{sec:appendix-user-simulators}

Each target language has its own locally served simulated-user model (Table~\ref{tab:model-access}), run with its default generation parameters. The model is prompted to reply in the target language and to reject any translation not in its language. The translated utterance is the user message and the simulated-user context is the system message. An overlong single-turn context is retried once after truncation to approximately 200 characters.

\begin{table*}[t]
  \centering
  \caption{Criteria per scenario and pooled pass rates (\%) by layer. $\tau$ is Kendall's rank correlation of the 10 setups with the row above it.}
  \label{tab:generator-layers}
  \small
  \begin{tabular}{@{}llccccccc@{}}
    \toprule
    & & \multicolumn{3}{c}{\textbf{Criteria per scenario}} & \multicolumn{3}{c}{\textbf{Pass rate (\%)}} & \\
    \cmidrule(lr){3-5}\cmidrule(lr){6-8}
    \textbf{Checklists} & \textbf{Judge} & \textbf{L1} & \textbf{L2} & \textbf{L3} & \textbf{L1} & \textbf{L2} & \textbf{L3} & \textbf{$\tau$} \\
    \midrule
    \multicolumn{9}{@{}l}{\textit{Without failure-point grounding}} \\
    Gemini 3.1 Pro, production prompt & Clef & 2.1 & 3.1 & 3.8 & 91.0 & 68.2 & 57.5 & -- \\
    Gemini 3.1 Pro, no failure-point grounding & Clef & 2.2 & 3.1 & 3.9 & 90.7 & 67.1 & 58.1 & 1.000 \\
    \midrule
    \multicolumn{9}{@{}l}{\textit{Independent generator}} \\
    Gemini 3.1 Pro & Clef & 2.9 & 6.3 & 7.6 & 83.9 & 48.5 & 33.0 & -- \\
    GPT-5.4 & Clef & 4.2 & 5.0 & 6.0 & 81.7 & 64.0 & 65.0 & 0.867 \\
    Gemini 3.1 Pro & Gemini 3.1 Pro & 2.9 & 6.3 & 7.6 & 79.8 & 53.3 & 37.5 & -- \\
    GPT-5.4 & Gemini 3.1 Pro & 4.2 & 5.0 & 6.0 & 81.3 & 70.7 & 69.8 & 0.911 \\
    \bottomrule
  \end{tabular}
\end{table*}

\subsection{User Simulation Validation}
\label{sec:appendix-simuser-validation}

We check the 60,000 single-turn simulated-user replies of the 10 setups. The refusal scan uses the final dataset, and the language and context checks use the replies from before the safety filter and the Bengali repair (Appendix~\ref{sec:appendix-bengali-repair}).

\textbf{Model choice.} Each simulated-user model is a monolingual or regional specialist. The simulated user only reads target-language text and replies in that language, so it needs no cross-lingual ability, and a specialist does not leak source-language text into a reply meant to come from a native speaker.

\textbf{Language fidelity.} GlotLID \citep{glotlid-2023}, with the same accepted dialect codes as \S\ref{sec:method-judge}, flags at most 3.7\% of Arabic, Bengali, and Korean replies in every setup. It flags 7.6--12.9\% of Indonesian replies because casual register falls just under the 0.9 confidence cutoff, although GlotLID still labels them Indonesian.

\textbf{Safety refusals.} A multilingual keyword scan finds replies that break character with a safety refusal (``my programming forbids...''), from 1.1--2.6\% of Indonesian turns to 6.1--11.9\% of Bengali turns across setups. Excluding these turns raises each setup's pass rate by 0.15 to 0.27 points and Bengali's by 0.95 points, so the refusals do not bias the pass rates.

\textbf{Context adherence.} The LaBSE similarity between each reply and its own conversation context exceeds a shuffled-context baseline in all 40 setup--language groups, by 0.14 to 0.21. Few replies are generic (0.5--8\%), and only SeamlessM4T v2 produces near-duplicate replies (1--1.5\%), which follow from its weaker translations. The simulated users therefore respond to their own context.

\subsection{Scoring Details}
\label{sec:appendix-scoring-details}

\paragraph{Failure handling.}
All setups keep the same 6,000-scenario denominator. 5 records have no usable translation (3 empty NLLB-200 outputs, and one Gemini 3.1 Pro and one SeamlessM4T v2 translation request that returned nothing) and fail every criterion with a pass rate of 0. No record is dropped.

\paragraph{Criteria-weighting sensitivity.}
\label{sec:appendix-criteria-weighting}
The pass rate (\S\ref{sec:method-judge}) counts every criterion equally. Because all setups are judged against the same checklist, we can also weight each criterion by how often the 10 setups fail it, or reweight by layer, by the Meaningfulness of its function (Appendix~\ref{sec:appendix-checklist-meaningfulness}), or by function. These scores use the criteria before LaBSE deduplication, so the equal-weight values differ slightly from Figure~\ref{fig:central-results}. Weighting by difficulty lowers all pass rates, since hard criteria count more, and only swaps Gemini 3.1 Flash Lite and DeepSeek V4 Pro, which stay within 2.2 points of each other. The other weightings keep the ranking of the 10 setups.

\paragraph{Generator without failure-point grounding.}
\label{sec:appendix-layer-ablation}
L3 is the only layer tied to a specific prompt requirement. Before writing criteria, the generator names two or three concrete ways the utterance could fail in translation because of the cultural gap between the two languages, its cultural failure points, and must ground at least one L3 criterion in them (Appendix~\ref{sec:appendix-prompts}). To check whether this requirement is what makes L3 hard, Gemini 3.1 Pro regenerated the checklists of 600 random scenarios (50 per direction) once with the production prompt and once without any reference to failure points. Clef judged all 10 setups against both versions on the 593 scenarios with valid checklists under both prompts. Removing the grounding barely changes the checklists or the pass rates (Table~\ref{tab:generator-layers}). L1 > L2 > L3 still holds for every setup, the ranking stays the same, and the gap between L1 and L3 changes by only $-0.8$ points (95\% CI $[-2.7, +1.1]$). The low L3 pass rates therefore come from the cultural-social criteria themselves.

\paragraph{Layers under an independent generator.}
\label{sec:appendix-generator-layers}
On the difficult scenarios of the generator test (\S\ref{sec:judge-validation-shared}) that have criteria in all three layers (465 under Clef and 462 under Gemini 3.1 Pro), we compare layer pass rates between the two checklist sets. With checklists written by GPT-5.4, L2 and L3 still stay below L1 for all 10 setups under both judges, but L2 and L3 become level (Table~\ref{tab:generator-layers}). GPT-5.4 writes fewer L3 criteria and phrases most of them as something to avoid (80\% against 53\% for Gemini 3.1 Pro), and its L3 criteria pass more often even when phrased positively (57.6\% against 30.1\% under Clef). How hard the cultural-social layer looks therefore depends on how strictly the generator writes it, while the gap between L1 and the two higher layers holds for both generators.

The two checklist sets give almost the same setup pass rates (Pearson $r=0.96$ under both judges), and all three rank changes under Clef involve Gemini 3.1 Flash Lite, which falls from 3rd to 6th. The cultural-adaptation brief asks for clarifications in brackets (Appendix~\ref{sec:appendix-prompts}). Flash Lite adds them to 99\% of its translations, against 4--32\% for the other LLMs, and GPT-5.4 writes more criteria against added content, so Flash Lite passes only 48\% of these criteria while the other LLMs pass 82--92\%.

\paragraph{Checklist size by layer.}
\label{sec:appendix-checklist-layer-distribution}

L3 makes up 43--46\% of the criteria in every target language and the most in Korean, whose honorific and register criteria add to L3 (Table~\ref{tab:layer-criteria-distribution}).

\begin{table}[htbp]
  \centering
  \caption{Mean criteria per scenario by layer and target language, pooled over the
  6{,}000-scenario dataset. L3 is 42.6--45.8\% of every language's checklist by count
  (overall 44.0\%, with the L1/L2 shares given in the text).}
  \label{tab:layer-criteria-distribution}
  \setlength{\tabcolsep}{4pt}
  \begin{tabular}{@{}lrrrr@{}}
    \toprule
    \textbf{Target} & \textbf{L1} & \textbf{L2} & \textbf{L3} & \textbf{Total} \\
    \midrule
    Arabic     & 2.94 & 6.02 & 6.94 & 15.90 \\
    Bengali    & 2.88 & 5.77 & 6.82 & 15.46 \\
    Indonesian & 3.00 & 6.60 & 7.13 & 16.72 \\
    Korean     & 2.87 & 6.02 & 7.51 & 16.41 \\
    \midrule
    Overall    & 2.92 & 6.10 & 7.10 & 16.12 \\
    \bottomrule
  \end{tabular}
\end{table}

%% file: appendix/D_judge_validation_detail.tex
\section{Complete Judge Validation}
\label{sec:appendix-judge-validation-detail}

\subsection{Judge Choice and Cross-Judge Agreement}
\label{sec:appendix-judge-choice}
\label{sec:judge-validation-choice}
We use Clef as the judge for three reasons. First, Gemini 3.1 Pro writes every checklist and is one of the interpreter setups, so as judge it would score its own checklists and translations. Second, Clef is open-weight and deterministic, so the results stay reproducible. Third, Clef runs locally at 2.2 seconds per judgment on four 11\,GB GPUs at no API cost, while Gemini 3.1 Pro costs USD 0.008 per judgment through the batch API, or USD 234.45 for the 28,862 multi-turn judgments. As checklist generator we use Gemini 3.1 Pro with high thinking effort, since annotators rate its criteria as meaningful (\S\ref{sec:checklist-validation}) and checklists regenerated by GPT-5.4 keep the ranking of the 10 setups (\S\ref{sec:judge-validation-shared}).

Clef receives the LLM judge prompt without the parts listed in Appendix~\ref{sec:appendix-clef}, which keeps it comparable to the LLM judges. On the 2,175 translations that all three LLM judges scored, Clef agrees with each of them about as well as they agree with each other (Cohen's $\kappa$ 0.42--0.58 against 0.43--0.58), and L3 is the lowest layer under every judge. Clef and Gemini 3.1 Pro also lead to the same conclusions in every experiment (Table~\ref{tab:clef-vs-llm}).
\input{tables/clef_vs_llm}

\subsection{Clef and Gemini 3.1 Pro on the Full Single-Turn Dataset}
\label{sec:appendix-clef-grid}
Over all 1.43 million judged criteria, the two judges agree on 82.2\% ($\kappa=0.62$), and Clef passes fewer criteria (60.9\% against 67.1\%). The two judges rank the 10 setups in the same order except for places 3--5 (Spearman $\rho=0.964$). Clef is stricter on the LLM interpreters than on dedicated MT, and it is more lenient only on SeamlessM4T v2 (Table~\ref{tab:clef-vs-gemini}).
\input{tables/clef_vs_gemini}

\subsection{Adversarial Judge Tests by Layer}
\label{sec:appendix-adversarial-detail}

\begin{figure}[htbp]
\centering
\includegraphics[width=\columnwidth]{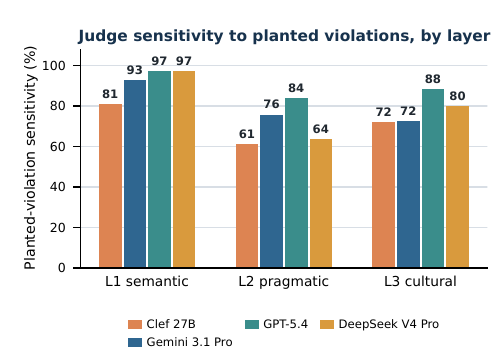}
\caption{Sensitivity to planted violations by judge and checklist layer (value on each bar).}
\label{fig:judge-sensitivity-appendix}
\end{figure}

Every judge detects fewer planted violations in L2 and L3 than in L1 (Figure~\ref{fig:judge-sensitivity-appendix}). Agreement between judges also drops from L1 to L3 for most judge pairs ($\kappa$ 0.45--0.61 on L1 and 0.40--0.56 on L3). The exceptions are GPT-5.4 paired with Gemini 3.1 Pro or with Clef, because GPT-5.4 passes fewer L1 criteria than the other judges (87\% against 94--95\%). The layers that are harder to preserve are therefore also harder to judge, and Clef's agreement with the annotators shows the same drop (Appendix~\ref{sec:appendix-human-calibration-detail}).

\subsection{Checklist Function Construction and Coherence}
\label{sec:appendix-function-construction}
Each scenario has its own checklist, so annotating single scenarios does not generalize. Following Evalet \citep{evalet-2025}, we pool every \texttt{(record, model, criterion)} instance the judge scored in each target language, embed them with LaBSE, and cluster them with spherical $k$-means separately per layer (10, 18, and 24 clusters for L1, L2, and L3, plus one residual cluster). An LLM labels each cluster as one yes/no question, a \emph{function}. This gives 53 clusters per target language, of which 39--41 remain active (Appendix~\ref{sec:appendix-function-results}). Mean cohesion within a function is 0.73--0.79. Human validation samples criteria by function (Appendix~\ref{sec:appendix-annotation-protocol}), with extra samples for about 14 loosely cohesive functions per language.

Functions are the only unit that recurs across records. A function covers a median of 461 records (up to 1,257), while flat scoring and per-scenario clustering give units that each cover a single record. Grouping criteria into functions moves the pass rate by only 2.0 points on average from flat scoring, against 5.4 points for per-scenario clustering, so functions make calibration across records possible without changing the scores. Judge disagreement within a function comes almost entirely from loosely cohesive clusters and is close to zero once members have cosine similarity of at least 0.85.

%% file: tables/clef_vs_llm.tex
\begin{table}[htbp]
  \centering
  \caption{Clef 27B against Gemini 3.1 Pro in each experiment: criterion-level $\kappa$ and AUROC, and one effect under each judge (points). Multi-turn $\kappa$ and AUROC cover L1--L3 criteria. $^\dagger$Flash Lite with history.}
  \label{tab:clef-vs-llm}
  \setlength{\tabcolsep}{2pt}
  \small
  \begin{tabular}{@{}lcclcc@{}}
    \toprule
    \textbf{Experiment} & \textbf{Kappa} & \textbf{AUC} & \textbf{Effect} & \textbf{Gem.} & \textbf{Clef} \\
    \midrule
    Single-turn & .62 & .91 & L1$>$L2$>$L3 & 10/10 & 10/10 \\
    Multi-turn (turn) & .60 & .90 & History & $+5.1$ & $+4.2$ \\
    Multi-turn (conv.) & .46 & .84 & Conv.$-$turn$^\dagger$ & $+8.3$ & $+3.8$ \\
    Ablation & .62 & .90 & Adapt.$-$spec. & $+2.3$ & $+2.9$ \\
    \bottomrule
  \end{tabular}
\end{table}

%% file: tables/clef_vs_gemini.tex
\begin{table}[htbp]
  \centering
  \caption{Canonical pass rate by judge (Clef at $\tau=0.5$).}
  \label{tab:clef-vs-gemini}
  \resizebox{\columnwidth}{!}{%
  \begin{tabular}{@{}lcccc@{}}
    \toprule
    \textbf{Setup} & \textbf{Gemini} & \textbf{Clef} & $\Delta$ & \textbf{Clef 95\% CI} \\
    \midrule
    Gemini 3.1 Pro          & 92.2 & 81.6 & $-10.6$ & [81.1, 82.0] \\
    DeepSeek V4 Pro         & 85.8 & 75.2 & $-10.6$ & [74.7, 75.7] \\
    Gemini 3.1 Flash Lite   & 80.5 & 74.0 & $-6.5$ & [73.5, 74.6] \\
    Qwen3.5 Flash           & 81.9 & 74.0 & $-7.9$ & [73.4, 74.5] \\
    GPT-5.4 Mini            & 80.9 & 71.7 & $-9.2$ & [71.2, 72.3] \\
    DeepSeek V4 Flash       & 79.6 & 70.3 & $-9.3$ & [69.7, 70.8] \\
    Google Translate        & 59.1 & 56.2 & $-2.9$ & [55.6, 56.8] \\
    NLLB-200 3.3B           & 56.8 & 53.0 & $-3.8$ & [52.4, 53.6] \\
    Tiny Aya                & 51.1 & 49.5 & $-1.6$ & [48.9, 50.1] \\
    SeamlessM4T v2 Large    & 27.1 & 32.3 & $+5.2$ & [31.7, 32.9] \\
    \bottomrule
  \end{tabular}}
\end{table}

%% file: appendix/E_human_calibration_detail.tex
\section{Human Validation Details}
\label{sec:appendix-human-annotation}

\subsection{Annotation Protocol}
\label{sec:appendix-annotation-protocol}

For each function, we sample translations the judge marked as met and as not met, spread evenly over source directions and the 10 interpreter setups, with extra samples for loosely cohesive functions (Appendix~\ref{sec:appendix-function-construction}).
Indonesian, Bengali, and Korean each have 3 native-speaker annotators, and Arabic has 2. Each annotator grew up in the culture of their language, uses it daily, and works fluently in English. One annotator per language also reviewed the pair-specific cultural context used by the checklist generator (\S\ref{sec:method-checklists}). Participants
were recruited through an open call circulated primarily within academic communities. The protocol received approval from an institutional review board, with details withheld for anonymous review.
Participation was voluntary, informed consent was documented through agreement and consent forms,
participants could withdraw at any time, and each annotator received 50,000 KRW for at most
five hours of work. The full annotation guideline is included in the supplementary material.

Each
per-target pool ($\sim$540--580 rows) splits into a shared overlap subset ($\sim$35\%, rated by
every annotator, giving Fleiss'/Cohen's $\kappa$) and a disjoint remainder split across
annotators. Every annotator completes two sheets. \textbf{Task A} (checklist quality) rates each
function's \emph{Meaningfulness} (1--5) and \emph{Answerability} (from target text + English gloss
alone, or needing the source) with no translations shown.
\textbf{Task B} (judge calibration) rates \emph{Satisfied} (yes/no) and \emph{Confidence} on
sampled exemplars, with model identity and the judge verdict hidden until after the annotator answers. 
The instructions require Task~A before Task~B, treat the reference as an aid, and reserve \emph{Unsure} for cases requiring the source.

\subsection{Checklist Meaningfulness}
\label{sec:appendix-checklist-meaningfulness}

Annotators use the 1--5 scale with different strictness, so their mean ratings differ widely within a language (Table~\ref{tab:meaningfulness}). We therefore compare annotators with Spearman correlation, which ignores these offsets. Korean has the lowest pooled mean because one annotator rates far more strictly than any other, while its Spearman range overlaps the other languages. Korean annotators also mark the most functions as needing the source.

\begin{table}[htbp]
  \centering
  \caption{Task~A ratings per target language. Mean Meaningfulness ratings are per annotator, Spearman is pairwise between annotators on the same 53 functions, and Source is the share of answerability ratings marked as needing the source.}
  \label{tab:meaningfulness}
  \small
  \setlength{\tabcolsep}{4pt}
  \begin{tabular}{@{}lccc@{}}
    \toprule
    \textbf{Target} & \textbf{Mean rating} & \textbf{Spearman} & \textbf{Source (\%)} \\
    \midrule
    Arabic & 3.91, 3.91 & 0.22 & 17.9 \\
    Bengali & 4.77, 3.51, 4.53 & 0.36--0.57 & 21.4 \\
    Indonesian & 3.26, 3.79, 4.89 & 0.34--0.48 & 22.8 \\
    Korean & 3.68, 3.23, 2.40 & 0.25--0.45 & 42.8 \\
    \bottomrule
  \end{tabular}
\end{table}

\subsection{Human--Judge Agreement by Language and Layer}
\label{sec:appendix-human-calibration-detail}

Clef's agreement with the annotators falls from L1 to L3 in every language (Table~\ref{tab:human-calibration}). GPT-5.4 and DeepSeek V4 Pro judged 3,364 and 3,033 of the 3,477 labels. On the 800 items labeled by two or three annotators, each judge agrees with the annotators within 0.4 points of its agreement on all labels.

\input{tables/human_calibration}

\textbf{Human--human agreement.} In the three languages with three annotators, annotators agree with each other on 68.6--77.2\% of the shared items, close to their agreement with the judges. Their $\kappa$ is much lower (Fleiss' $\kappa$ 0.09--0.28) because about 80\% of the labels are yes, so two annotators already agree about two thirds of the time by chance.

\textbf{Why agreement is not higher.} Labels that annotators mark as Unsure agree less with Clef in Bengali, Indonesian, and Korean (Table~\ref{tab:sure-unsure}), which fits annotators who cannot check the translation against the source. Arabic is the exception, where Unsure labels agree slightly more. Functions that annotators rate as needing the source in Task~A also agree less in Indonesian and Korean, but not in Bengali and Arabic.

\begin{table}[htbp]
  \centering
  \caption{Agreement (\%) with the annotators by their stated confidence. Unsure is the share of labels marked Unsure.}
  \label{tab:sure-unsure}
  \small
  \setlength{\tabcolsep}{4pt}
  \begin{tabular}{@{}lccccc@{}}
    \toprule
    & & \multicolumn{2}{c}{\textbf{Clef}} & \multicolumn{2}{c}{\textbf{Gemini 3.1 Pro}} \\
    \cmidrule(lr){3-4}\cmidrule(lr){5-6}
    \textbf{Target} & \textbf{Unsure} & \textbf{Sure} & \textbf{Unsure} & \textbf{Sure} & \textbf{Unsure} \\
    \midrule
    Arabic & 17.0 & 75.7 & 79.2 & 88.2 & 89.6 \\
    Bengali & 13.7 & 77.4 & 49.6 & 74.8 & 58.8 \\
    Indonesian & 8.4 & 69.6 & 53.6 & 70.4 & 61.9 \\
    Korean & 11.3 & 74.0 & 54.4 & 77.3 & 63.1 \\
    \midrule
    All & 12.2 & 73.9 & 59.0 & 76.4 & 68.2 \\
    \bottomrule
  \end{tabular}
\end{table}

\subsection{Cascade to the Full Dataset}
\label{sec:appendix-cascade}
\input{tables/cascade_shift}

The calibration design propagates each \texttt{(function, Clef verdict)} agreement rate to \emph{every}
instance of that function dataset-wide, including scenarios no annotator ever read, as
$\hat p_{\text{true met}} = a(f,\text{met})$ if the judge said met, else $1-a(f,\neg\text{met})$,
with an empirical-Bayes Beta-Binomial fit shrinking small-$n$ functions toward their layer mean.

\begin{figure}[htbp]
  \centering
  \includegraphics[width=\columnwidth]{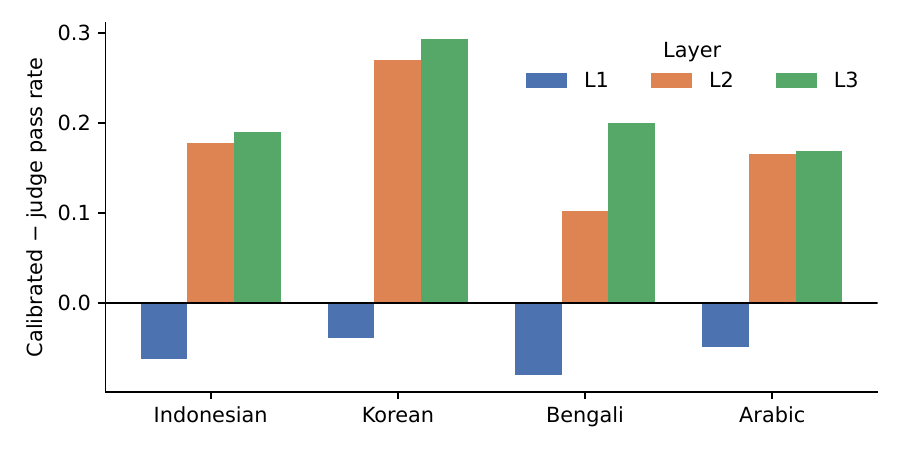}
  \caption{Change in Clef's pass rate after calibration to the annotators, by target language and layer.}
  \label{fig:cascade-layer-shift}
\end{figure}

Calibration changes the layer profile without changing the ranking (Figure~\ref{fig:cascade-layer-shift}). After calibration L1 and L2 are level, L3 stays the lowest layer for 9 of the 10 setups (SeamlessM4T v2 is the exception), and the ranking of the 10 setups is unchanged (Kendall $\tau=1$).

The calibration assumes that the judge's agreement with the annotators is a property of the function. To test this, we predict each annotated label from the other labels of its function. The mean absolute error is 0.22--0.27 per language, below the 0.30--0.32 obtained on simulated labels for which the assumption holds by construction, so the function is a sound unit for calibration.

\begin{figure*}[t]
\centering
\includegraphics[width=0.78\textwidth]{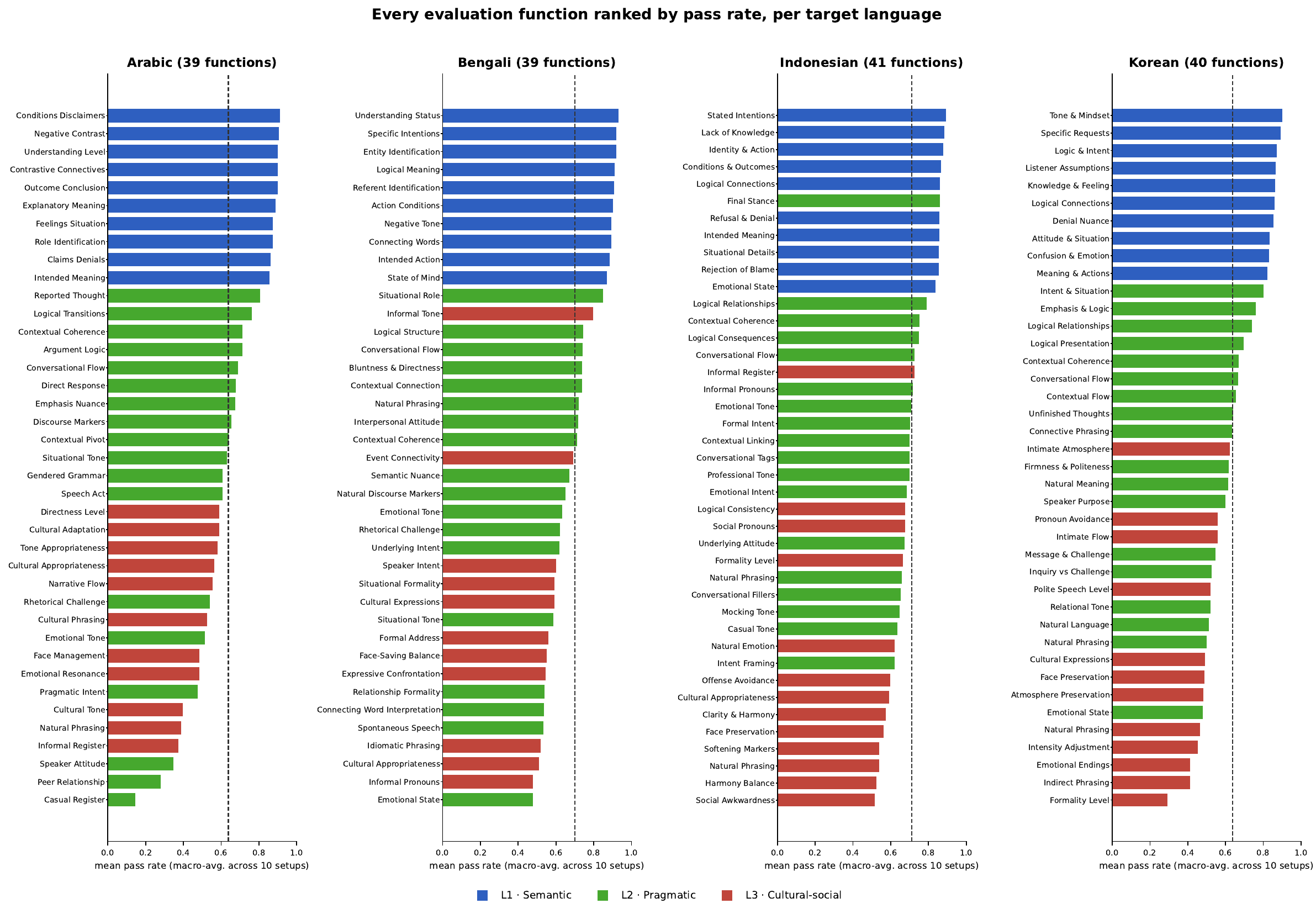}
\caption{Every function's pass rate averaged over the 10 setups, ranked within each target language and colored
by layer. The dashed line marks the language's mean.}
\label{fig:function-ranked-bars}
\end{figure*}

%% file: tables/human_calibration.tex
\begin{table}[htbp]
  \centering
  \caption{Human--judge agreement (\%) for Clef and Gemini 3.1 Pro on the 3,477 Task~B labels that map to a judged criterion. Some exported records
  use an unrecognized layer label, so layer counts need not sum to the overall count.}
  \label{tab:human-calibration}
  \resizebox{\ifdim\width>\linewidth\linewidth\else\width\fi}{!}{%
  \begin{tabular}{@{}llcccc@{}}
    \toprule
    \textbf{Target lang.} & \textbf{Layer} & \textbf{$n$} & \textbf{Clef} & \textbf{Gemini} & \textbf{Annotators} \\
    \midrule
    Arabic & L1      & 104 & 89.4 & 88.5 & \multirow{4}{*}{2} \\
    Arabic & L2      & 193 & 75.1 & 89.6 &  \\
    Arabic & L3      & 308 & 72.1 & 88.3 &  \\
    Arabic & Overall & 624 & 76.3 & 88.5 &  \\
    \addlinespace
    Bengali & L1      & 177 & 83.6 & 84.7 & \multirow{4}{*}{3} \\
    Bengali & L2      & 283 & 70.3 & 69.6 &  \\
    Bengali & L3      & 462 & 71.0 & 69.5 &  \\
    Bengali & Overall & 946 & 73.6 & 72.6 &  \\
    \addlinespace
    Indonesian & L1      & 195 & 76.9 & 75.9 & \multirow{4}{*}{3} \\
    Indonesian & L2      & 302 & 72.8 & 69.2 &  \\
    Indonesian & L3      & 460 & 60.9 & 67.0 &  \\
    Indonesian & Overall & 995 & 68.2 & 69.6 &  \\
    \addlinespace
    Korean & L1      & 171 & 77.2 & 72.5 & \multirow{4}{*}{3} \\
    Korean & L2      & 286 & 69.9 & 72.7 &  \\
    Korean & L3      & 421 & 70.3 & 78.9 &  \\
    Korean & Overall & 912 & 71.8 & 75.7 &  \\
    \bottomrule
  \end{tabular}
  }
\end{table}

%% file: tables/cascade_shift.tex
\begin{table}[t]
  \centering
  \caption{Calibration of Clef to the target-only annotators: function--verdict cells with at least two annotations, Clef pass rate, and calibrated pass rate.}
  \resizebox{\columnwidth}{!}{%
  \begin{tabular}{@{}lccc@{}}
    \toprule
    \textbf{Lang.} & \textbf{Cells (n$\geq$2)} & \textbf{Clef} & \textbf{Calib.\ ($\Delta$)} \\
    \midrule
    Indonesian & 99/106 (93\%) & 0.658 & 0.813 (+0.155) \\
    Korean     & 97/106 (92\%)   & 0.561 & 0.809 (+0.247) \\
    Bengali    & 90/106 (85\%)   & 0.640 & 0.779 (+0.139) \\
    Arabic     & 86/106 (81\%)   & 0.581 & 0.722 (+0.141) \\
    \bottomrule
  \end{tabular}}

  \label{tab:cascade-shift}
\end{table}

%% file: appendix/F_language_direction_results.tex
\section{Language, LID, and Function-Level Results}
\label{sec:appendix-language-results}

\subsection{Target- and Source-Language Effects}
\label{sec:appendix-stats-lid-lang}
\label{sec:appendix-stats-source-lang}
We run one Friedman test per role over the 10 setups' mean pass rate in each language. The language effect is large in both roles (target $\chi^2(3)=21.96$, Kendall's $W=0.73$; source $\chi^2(3)=22.56$, $W=0.75$; both $p<0.001$), so the setups largely agree on which languages are hard. Compared with each setup's own mean, Indonesian is easier and Korean harder as target in all 10 setups. As source, Arabic, Indonesian, and Korean differ significantly from the setup mean while Bengali does not. Arabic is harder as target and easier as source, which points to a problem of adapting into Arabic. Under Gemini 3.1 Pro as judge, Korean is also the hardest and Indonesian the easiest language in both roles.

\subsection{LID Acceptance and Filter}
\label{sec:appendix-lid}
GlotLID \citep{glotlid-2023} runs on every translation. Because automatic LID can
misidentify a correct translation as a related dialect or macrolanguage, an exact-code
match is not required for every target. For an Arabic target, detection is accepted if it lands in
a curated set of ISO 639-3 codes covering the Arabic macrolanguage, Modern Standard Arabic, and its
regional dialects (Egyptian, Levantine, Gulf, Moroccan, Najdi, Iraqi, Tunisian, and others). For an
Indonesian target, a ``Malay-cluster'' set (standard Malay, Betawi, and several Malay varieties) is
accepted outright, since these are mutually intelligible with Indonesian.
Bengali and Korean targets use a single-code match, since GlotLID's label set has no such dialect ambiguity for them.

The main results do not filter by LID. The strict filter, which gives a pass rate of zero to every translation LID rejects, lowers each setup's pass rate by 0.9 to 2.1 points, significant for every setup. The drop depends on both how often a setup is rejected and how well its rejected translations would have scored, so SeamlessM4T v2, rejected most often, loses only 1.1 points. Most of the drop falls on Indonesian targets (4.9 points against 0.1--0.6 for the other languages), where GlotLID's confidence threshold rejects many correct translations. The hatched end of each bar in Figure~\ref{fig:central-results} shows the pass rate each setup loses to the filter.

\subsection{Function-Level Results}
\label{sec:appendix-function-results}
To show what the layer averages hide, we pool the criteria of each target language by function. For each function we compute its pass rate within each setup and average it over the 10 setups, which gives pass rates for 159 active functions. Figure~\ref{fig:function-ranked-bars} ranks the 39--41 functions of each target language. Weak functions differ by language, such as casual register in Arabic and speech-level formality and emotion-marking endings in Korean, so the same layer can fail for different reasons in each language. Function labels are short descriptions.

%% file: appendix/G_multiturn_study.tex
\section{Multi-Turn Conversation Study}
\label{sec:appendix-multiturn-pilot}
\input{tables/prompt_ablation_full}
\subsection{Real Dialogue Construction}
Each conversation is a window of 13 subtitle lines from the segment pool of \S\ref{sec:data-construction}. Consecutive lines of one speaker are merged into one turn, since strict alternation occurs in only about 1\% of windows, and the first six turns are kept. Every Bengali line in a window must pass the vowel-sign check of Appendix~\ref{sec:appendix-bengali-repair}. Gemini 3.1 Flash Lite keeps windows with two speakers and aligned turns, Gemini 3.8 Flash labels the speakers, and we keep 100 conversations per pair. Each conversation gets only a neutral context line (``A conversation between two people, from a film'').

Turn-level checklists use the single-turn template, with the 15 preceding subtitle lines as context and the other language's line as reference. GPT-5.4 wrote the 25 checklists that Gemini 3.1 Pro failed to produce. For ten Qwen3.5 Flash judgments that exceeded the judge's GPU memory, we cut returned reasoning to 300 characters and earlier turns to 600 (1,000 at conversation level).

\subsection{Scripted Mode by Pair}
\label{sec:appendix-multiturn-scripted}
Flash Lite with history leads in every pair (75.1--80.1\%). NLLB-200 and Tiny Aya are weakest on the three Korean pairs (47.1--55.6\%) and strongest on Arabic--Bengali, Arabic--Indonesian, and Bengali--Indonesian (60.9--70.0\%).

\subsection{Dynamic Mode}
\label{sec:appendix-multiturn-dynamic}
\begin{figure}[htbp]
  \centering
  \includegraphics[width=\columnwidth]{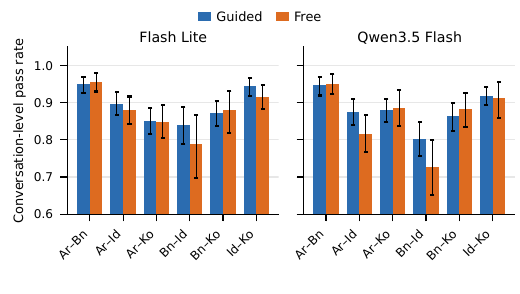}
  \caption{Dynamic conversations, conversation-level pass rate by pair, guided and free (Clef judge, 95\% bootstrap CI).}
  \label{fig:multiturn-guided-vs-free-main}
\end{figure}

Dynamic conversations are produced live by the simulated users (\S\ref{sec:method-multiturn}) and scored with
the single-turn template without a reference translation. Arabic simulated-user turns ended with a leaked
end-of-turn token, which we strip before checklist generation and judging (1,572 turns).
Free conversations score higher than guided ones at conversation level in 5 of the 12 setup--pair combinations (Figure~\ref{fig:multiturn-guided-vs-free-main}).
The largest guided--free differences are for Qwen3.5 Flash on Bengali--Indonesian, where free conversations score 7.5 points below guided ones, and on Arabic--Indonesian (5.9 points below), and neither difference is significant.

\subsection{Clef and Gemini 3.1 Pro on Multi-Turn}
Under Gemini 3.1 Pro as judge, the setups keep the same order as under Clef. Against Gemini's own single-turn pass rates, turn-level pass rates change by $-1.9$ to $+3.2$ points for the LLM setups and rise by 6.7 and 6.6 points for NLLB-200 and Tiny Aya, as under Clef. Agreement with Clef and the history and conversation-level effects are in Table~\ref{tab:clef-vs-llm}.

%% file: tables/prompt_ablation_full.tex
\begin{table*}[t]
  \centering
  \caption{Pass rate (\%) of the seven setups under the six briefs on 1{,}108--1{,}113 paired difficult scenarios per setup. Bold marks each row's best brief.}
  \label{tab:prompt-ablation-full}
  \setlength{\tabcolsep}{6pt}
  \small
  \begin{tabular*}{\textwidth}{@{\extracolsep{\fill}}lcccccc@{}}
    \toprule
    \textbf{Setup} & \textbf{\shortstack{Literal,\\no context}} & \textbf{\shortstack{Literal\\+ context}} & \textbf{\shortstack{Neutral,\\no context}} & \textbf{\shortstack{Neutral\\+ context}} & \textbf{\shortstack{Specification-\\aware}}
      & \textbf{\shortstack{Cultural\\adaptation}} \\
    \midrule
    Gemini 3.1 Pro & 47.1 & 48.9 & 54.2 & 61.0 & 60.1 & \textbf{67.9} \\
    DeepSeek V4 Pro & 47.8 & 49.9 & 51.6 & 56.5 & 56.9 & \textbf{61.4} \\
    GPT-5.4 Mini & 48.9 & 51.4 & 50.6 & 54.2 & 55.2 & \textbf{56.8} \\
    DeepSeek V4 Flash & 45.7 & 47.9 & 48.6 & 50.9 & 53.0 & \textbf{55.2} \\
    Gemini 3.1 Flash Lite & 48.4 & 50.8 & 52.2 & 56.9 & \textbf{57.3} & 54.6 \\
    Qwen3.5 Flash & 43.4 & 45.4 & 48.8 & 52.9 & 52.5 & \textbf{54.4} \\
    Tiny Aya & 35.9 & 38.4 & 39.0 & \textbf{41.0} & 38.1 & 38.7 \\
    \bottomrule
  \end{tabular*}
\end{table*}

%% file: appendix/H_prompt_ablation.tex
\section{Prompt Ablation, Metric Correlations, and Judging Without a Checklist}
\label{sec:appendix-prompt-ablation}

\subsection{Prompt-Ablation Details}

The ablation evaluates seven setups on 1,108--1,113 paired difficult scenarios under six interpreter briefs.
It changes only the translation brief and available context. Checklist generation and judging
remain fixed. Each output is scored like the main single-turn results, as the share of the scenario's deduplicated L1--L3 criteria met, without the LID filter. The two neutral briefs replace the literal-translation instruction with ``Translate the message from one language to the other.'' and are the shortest briefs (Appendix~\ref{sec:appendix-prompts}). With context, the neutral brief still matches the much longer specification-aware brief, also under Gemini 3.1 Pro as judge, so brief length does not explain the gains.

The capability-dependent benefit described in \S\ref{sec:analysis-prompting} holds for all seven setups across the literal, specification-aware, and cultural-adaptation briefs in Table~\ref{tab:prompt-ablation-full}, with one exception: Gemini 3.1 Flash Lite scores
2.7 points higher under the specification-aware brief than under the cultural-adaptation brief, a difference that is significant (95\% CI $[+1.6, +3.9]$). The difficult subset holds the scenarios on which Gemini 3.1 Flash Lite scored lowest under the
cultural-adaptation brief, so selection lowers its cultural-adaptation pass rate in particular. Its per-setup comparison therefore
contains a selection effect and should not be interpreted as a clean population estimate. The
pooled comparison in the main text is retained as an intervention study.

\subsection{Power of the MT-Metric Correlations}
\label{sec:appendix-stats-power}

Mixing all 10 setups' outputs together ($n\approx60{,}000$) gives Spearman $\rho$ of 0.28 for CometKiwi, 0.36 for GEMBA-MQM, 0.29 for MQM-APE, and 0.44 for RATE (95\% CI within $\pm0.01$). Each of these exceeds every within-setup value for the two strongest interpreters, so the pooled correlation comes almost entirely from metrics ranking weak MT baselines below strong LLM interpreters.

\subsection{Judging Without a Checklist}
\label{sec:appendix-holistic}

To test whether the scenario context alone explains the gap between our checklist and the MT metrics, Clef judged
every translation with the same input as the main grid (scenario context, source, translation, recipient reply, and
language checks) and two holistic yes/no questions in place of the checklist: ``The translation lets the recipient
understand what the speaker meant, with the same intent and tone, in a way that is natural and socially appropriate
for this conversation'' (H1) and ``This is a good translation of the source message for this conversation'' (H2).
Seven of Gemini 3.1 Pro's translations exceed the judge's input length, which leaves 5,992.

On these, H1 passes 94.4\% and H2 93.5\%. H1 correlates with CometKiwi at Spearman $\rho=0.34$, against 0.11 for our
checklist pass rate, and with GEMBA-MQM, MQM-APE, and RATE at 0.14--0.19, against 0.00--0.05. Of the 243 translations that
CometKiwi places in its top quartile and the checklist pass rate is at most 0.6, H1 fails 10.7\% (H2 10.3\%), against 1.7\%
of the top-quartile translations that pass the checklist. With the same context and no checklist, the judge therefore
behaves like a lenient fidelity judge. Across all 10 setups (59,935 translations), H1 ranks the setups with Kendall
$\tau=0.78$ against the checklist ranking (H2 0.73), and it fails 19.0\% of the 4,705 translations that CometKiwi
places in its top quartile and the checklist fails.

%% file: appendix/I_full_prompts.tex
\section{Full Prompts}
\label{sec:appendix-prompts}

These templates follow the supplementary code. Placeholders appear in \texttt{\{braces\}}, and lines beginning
\texttt{[EDITORIAL]} distinguish combined variants and were not sent to a model. Pair-specific
cultural context and function taxonomies remain as placeholders. Unsupported non-Latin examples
use Unicode escapes, and Unicode punctuation and symbols (dashes, arrows, bullets, $\leq$) are rendered in ASCII.

\newtcblisting{PromptBox}[1]{
  breakable,
  enhanced,
  top=3pt,
  left=4pt,
  right=4pt,
  bottom=3pt,
  title={#1},
  title after break={#1 (continued)},
  before skip=10pt,
  after skip=10pt,
  listing only,
  listing options={
    basicstyle=\ttfamily\scriptsize,
    breaklines=true,
    breakatwhitespace=false,
    columns=fullflexible,
    keepspaces=true,
    showstringspaces=false
  }
}

\subsection{Interpreter Translation Prompts}

The \emph{cultural-adaptation} condition (\texttt{cultural\_context} in the code) uses the production brief below. It receives the scenario context and no pair-specific cultural context. Specification-aware, neutral, and
literal conditions replace only that system instruction, while multi-turn combines the production brief
with the multi-turn task variant.

\begin{PromptBox}{System Prompt: Cultural-Adaptation Brief}
You are an expert translator and interpreter facilitating communication between two users.
- User A Language: {user_a_language}
- User B Language: {user_b_language}
- Conversation Context: {conversation_context}

Note: You are only provided with the languages of the users. Do not assume any additional user background.

## Core Instructions
1. **Sole Intermediary**: The users communicate exclusively through you. You are their only bridge.
2. **Liberal Adaptation**: You are encouraged to translate liberally to achieve naturalness and cultural relevance. Do not translate literally (word-for-word). Your priority is to convey the *intent* and *impact* of the message, limiting structural changes to what is necessary for naturalness.
3. **Explicate the Implicit**: If the source text contains implicit cultural context (e.g., social hierarchy, religious norms, gender distinctions) that is critical for the target user to understand, you must make it clear.
4. **Preserve Communicative Goal**: While the phrasing should be adapted, the core message and the speaker's intent must remain faithful to the source.

## Guidelines
1. **Necessary Adaptation**: Translate the situation, not just the words. Adapt idioms, honorifics, and cultural references to feel native to the target user, but ensure adaptations form a bridge, not a barrier. Do not over-localize.
2. **Contextual Clarity**: If a concept in the source language implies specific needs or rules that are not obvious in the target language, you must clarify them.
3. **Tone and Style**: Adjust the tone (formal/casual) to match the target culture's norms for the given situation.
4. **Bracketed Clarifications**: Any additional clarification or context needed for understanding MUST be placed inside brackets () and written in the **target language** -- never in English or any other language. Do not add translator's notes, headings, or meta-commentary in any language other than the target language.

## Quality Standards
- **Naturalness**: The translation should sound like it was originally spoken in the target language.
- **Cultural Intelligence**: The target user should understand the full implication of the message.
- **Faithfulness**: The underlying intent of the speaker is preserved.
\end{PromptBox}

\begin{PromptBox}{System Prompt: Specification-Aware Brief}
You are a professional translator working from an explicit translation specification (Kayano & Sugawara, 2025).
- User A Language: {user_a_language}
- User B Language: {user_b_language}
- Conversation Context: {conversation_context}

## Translation Specification
1. **Purpose of Translation**: The communicative goal of the message -- the two users share no common language and communicate solely through your translation.
2. **Target Audience**: The intended reader's language background and expectations -- a {user_b_language} speaker, with no assumed background beyond what the conversation context provides.
3. **Style, Register, and Tone**: The formality, style, and tone appropriate for the target context; match the source message's register as closely as the target language permits.
4. **Terminology and Reference Resources**: Preserve names, numbers, and domain-specific terms exactly, consistent with prior usage in the conversation.
5. **Domain and Legal Requirements**: Follow the norms appropriate to the message's domain and any applicable compliance considerations.
6. **Cultural Adaptation**: Make adjustments for cultural norms or sensitivities where appropriate.
7. **Length and Formatting**: Constraints on text length, layout, or structure -- keep the translation close to the source in length and structure.
8. **Localization Needs**: Regional or language-variant customization appropriate to the target audience.

Output ONLY the translation in {user_b_language}.
\end{PromptBox}

The two literal conditions are identical except for the marked context line.

\begin{PromptBox}{System Prompt: Literal Brief (With and Without Context)}
You are a translator facilitating communication between two users.
- User A Language: {user_a_language}
- User B Language: {user_b_language}
[EDITORIAL -- literal-with-context only: - Conversation Context: {conversation_context}]

Translate the message directly and literally from one language to the other. Do not add clarifications, adapt tone or register, or localize cultural references -- preserve the source phrasing as closely as target-language grammar allows.
\end{PromptBox}

The two neutral conditions use the literal brief's first lines and replace its final instruction.

\begin{PromptBox}{System Prompt: Neutral Brief (With and Without Context)}
You are a translator facilitating communication between two users.
- User A Language: {user_a_language}
- User B Language: {user_b_language}
[EDITORIAL -- neutral-with-context only: - Conversation Context: {conversation_context}]

Translate the message from one language to the other.
\end{PromptBox}

The output instruction is shared. The marked phrases and transcript block appear only in the
multi-turn task.

\begin{PromptBox}{User Prompt: Translation Task (Single- and Multi-Turn)}
[EDITORIAL -- choose the applicable opening line]
[SINGLE-TURN] Task: Translate the following message from {from_language} to {to_language}.
[MULTI-TURN] Task: Translate the following message from {from_language} to {to_language}, as part of an ongoing two-party conversation mediated by you, the interpreter.
{context}
[EDITORIAL -- multi-turn with history only: {transcript_block} = "Conversation so far:" and one line per earlier turn, "Turn {turn_index} ({speaker}): {source_text} | this run's translation: {translation}"]
[EDITORIAL -- choose the applicable message line]
[SINGLE-TURN] Message to translate: {message}
[MULTI-TURN] Message to translate ({speaker}, turn {turn_index}): {message}

Output ONLY the translation in {to_language}. Any bracketed clarifications must also be in {to_language}. Do not include English headings, notes, or meta-commentary.

Translation:
\end{PromptBox}

\subsection{Single-Turn Data Augmentation and Checklist Generation}

Single-turn augmentation and checklist generation are one implemented call. The prompt below
constructs pragmatic metadata and role-play contexts as well as the three checklist layers and the
flattened verification prompt.
The \texttt{\{cultural\_context\_block\}} placeholder holds the pair-specific cultural context (\S\ref{sec:method-checklists}). The six documents are released with the benchmark.

\begin{PromptBox}{Single-Turn Data Augmentation and Checklist Generation Prompt}
You are an expert evaluation data designer for cross-cultural interpreter-mediated communication.

Your task: convert a bilingual subtitle segment into structured evaluation metadata for a one-turn interpretation simulation.

Direction: {source_language} ({source_language_code}) -> {target_language} ({target_language_code})

{cultural_context_block}Source turn (what the interpreter receives):
{source_text}

Reference target turn (pragmatic reference only -- not ground truth):
{reference_target_text}
{reference_alignment_note}

Detected linguistic features and their evaluation implications:
{reason_tags_block}

Context digest (must be reflected in outputs):
{context_digest}

Previous context window:
{prev_context}

====================================================
STEP 1 -- Pragmatic analysis (reason before generating)
====================================================

Analyze the source turn and produce a compact analysis covering ALL of:
A. Speech act: primary communicative act (request / refusal / apology / assertion / question / complaint / promise / greeting / challenge / other)
B. Social relationship: power/solidarity dynamic presupposed (superior->subordinate / peer / subordinate->superior / stranger / intimate / etc.)
C. Face stakes: is there a face-threatening act? What mitigation does the source culture use, and what does the target culture expect instead?
D. Cultural failure points: given the pair-specific asymmetries above, name 2-3 concrete ways this specific utterance could fail in translation -- grounded in the actual cultural gap, not generic errors.
E. Required target form: what register, honorific level, and grammatical form must the target use?

Output this as "pragmatic_analysis": a 3-5 sentence paragraph that a human judge could use to evaluate the translation.

====================================================
STEP 2 -- Generate all output fields
====================================================

Using your Step 1 analysis, generate all fields below.

CHECKLIST FORMAT RULE -- mandatory for every item:
Every checklist item must be a yes/no question starting with "Does the" targeting:
- "Does the translation ..." -- Layer 1: fidelity and semantic accuracy
- "Does the interpreter's response ..." -- Layers 2 & 3: pragmatic function and communicative goal

The distinction matters: Layer 1 checks whether the translation is *accurate*. Layers 2-3 check whether the interpreter's *response achieves its communicative goal* in the target cultural context -- which may require culturally-adapted choices beyond word-for-word accuracy.

TASK REQUIREMENTS:
1) Infer speech_act_intent (<=8 words) and semantic_core from source + context.
2) Produce mandatory_cultural_constraints grounded in Step 1 cultural failure points (D).
3) Build roleplay-ready contexts while de-identifying movie specifics.
4) Do not include actor names, film titles, or scene-specific lore.
5) user_a_context must be in source language. user_b_context must be in target language.
6) Checklist items must use the "Does the translation/interpreter's response ..." format (YES = success).
7) Enforce checklist priority: layer_3 count >= layer_2 count >= layer_1 count.
8) conversation_context must be grounded in previous context window only; no transcript-style turn history.
9) Do not include the current source turn in conversation_context, context_window_summary, or user contexts.
10) context_window_summary: 2-4 English sentences about previous context only.
11) Both user contexts: rich role/situation grounding in each user's language without exposing past utterances.
12) user_a_context must explicitly identify User A (source-side); user_b_context must identify User B (target-side).
13) Do not include guidance phrasing: "Saya harus/akan", "I must/will", "\uD574\uC57C", "\uD560 \uAC70".
14) Do not include target-side plans, expected replies, or strategy hints in user_b_context.
15) verification_prompt: numbered lines (1., 2., ...).
16) Checklist must include at least one criterion for contextual coherence with surrounding turns.
17) layer_3 must include at least one criterion grounded in the Step 1 cultural failure points (D).

Output JSON only with this schema:
{
  "pragmatic_analysis": "string",
  "speech_act_intent": "string",
  "semantic_core": "string",
  "mandatory_cultural_constraints": ["string"],
  "context_window_summary": "string",
  "conversation_context": "string",
  "user_a_context": "string",
  "user_b_context": "string",
  "checklist": {
    "layer_1_semantic_core": ["Does the translation ..."],
    "layer_2_pragmatic_function": ["Does the interpreter's response ..."],
    "layer_3_cultural_social_constraints": ["Does the interpreter's response ..."]
  },
  "verification_prompt": "string"
}
\end{PromptBox}

\subsection{Multi-Turn Checklist Generation}
Turn-level checklists use the single-turn template above. For scripted turns, the previous context is the
15 subtitle lines before the turn and the reference is the other language's subtitle line. Dynamic turns have no
reference: the previous context is the earlier turns, and the reference line reads ``(no reference translation is
available for this turn)''. The conversation-level checklist applies the edits below to the same template.
\begin{PromptBox}{Conversation-Level Template Edits}
[EDITORIAL -- each line: original text => replacement]
[EDITORIAL -- dynamic conversations: the reference block reads "(no reference translation is available for this conversation)"]
Your task: convert a bilingual subtitle segment into structured evaluation metadata for a one-turn interpretation simulation. => Your task: convert a bilingual {n_turns}-turn conversation into structured evaluation metadata for a whole-conversation interpretation simulation, in which the interpreter translates every turn.
Direction: {source_language} ({source_language_code}) -> {target_language} ({target_language_code}) => Speakers: User A speaks {source_language} ({source_language_code}); User B speaks {target_language} ({target_language_code}). The interpreter translates each turn into the other speaker's language.
Source turn (what the interpreter receives): => Conversation (what the interpreter receives; each turn in its speaker's language):
Reference target turn (pragmatic reference only -- not ground truth): => Reference translation of each turn (pragmatic reference only -- not ground truth):
Analyze the source turn and produce a compact analysis covering ALL of: => Analyze the conversation as a whole (all turns together) and produce a compact analysis covering ALL of:
name 2-3 concrete ways this specific utterance could fail in translation => name 2-3 concrete ways this specific conversation could fail in translation, including failures that only show across turns (register or honorific drift, inconsistent rendering of names or terms, lost references to earlier turns)
9)  Do not include the current source turn in conversation_context => 9)  Do not include any turn of the conversation in conversation_context
17) layer_3 must include at least one criterion grounded in the Step 1 cultural failure points (D). => 17) layer_3 must include at least one criterion grounded in the Step 1 cultural failure points (D). [new line] 18) Every checklist item must judge the translated conversation as a whole across its turns, not any single turn: consistency of register and honorifics, consistent rendering of names and terms, coherent flow and references between turns, and whether the exchange keeps its overall communicative purpose.
\end{PromptBox}
\subsection{Judge Prompt}

\begin{PromptBox}{Judge Prompt}
You are an expert linguistic and cultural evaluator.
Your task is to evaluate the quality of a translation given the conversation context, source text, the translation, and the target recipient's response.

Conversation Context: {conversation_context}
Source Text: "{source_text}"
Translated Text: "{translated_text}"
Target Recipient Response: "{target_response}"

Language Verification Results:
{language_verification_info}

Verification Checklist:
{verification_prompt}

For each item in the verification checklist, determine if the translation successfully meets the criteria (Yes/No).
Also provide a brief reasoning for your decision.

IMPORTANT: If the language verification indicates that the target recipient's response is in the wrong language, this typically means the communication has failed. Any criteria that depend on the appropriateness or correctness of the target's response should likely be marked as "not met" since responding in the wrong language is a critical failure.

Evaluate each criterion carefully based on:
1. Translation Accuracy: Did the interpreter correctly handle the linguistic/cultural challenge?
2. Pragmatic Outcome: Did the communication succeed based on User B's response (considering language issues)?

Format your output as a JSON object with a "results" array only (do not include completion_rate).
Each result should have: id (number), criteria (string), met (boolean), and reasoning (string).
\end{PromptBox}

Clef receives this prompt without the verification checklist block, the reasoning request, and the two output-format lines; each criterion is passed as one schema field (Appendix~\ref{sec:appendix-clef}).

\subsection{Multi-Turn Judge Prompt}
Each multi-turn turn is judged with the prompt below, which adapts the single-turn Judge Prompt to one turn of a conversation: it adds the preceding turns, the turn index and speaker, and a consistency criterion. The conversation-level judgment uses a separate prompt, given after it, that takes the whole transcript and the conversation-level checklist. Clef receives both prompts without the checklist block, the reasoning request, and the output-format lines (Appendix~\ref{sec:appendix-clef}).
\begin{PromptBox}{Multi-Turn Judge Prompt (Turn)}
You are an expert linguistic and cultural evaluator, judging one turn of an ongoing interpreted two-party conversation.

Conversation Context: {conversation_context}

Conversation so far (turns before this one):
{transcript_block}

This turn (turn {turn_index}, speaker {speaker}):
Source Text: "{source_text}"
Translated Text: "{translated_text}"
Listener Response: "{listener_response}"

Language Verification Results:
{language_verification_info}

Verification Checklist (for THIS turn only):
{verification_prompt}

For each item in the verification checklist, determine if the translation successfully meets the criteria (Yes/No).
Also provide a brief reasoning for your decision.

IMPORTANT: If the language verification indicates that the translation or the listener's response is in the wrong
language, this typically means the communication has failed. Any criteria that depend on the appropriateness or
correctness of the translation/response should likely be marked as "not met" since responding in the wrong language
is a critical failure.

Evaluate each criterion carefully based on:
1. Translation Accuracy: Did the interpreter correctly handle the linguistic/cultural challenge in THIS turn?
2. Pragmatic Outcome: Did the communication succeed based on the listener's response (considering language issues) and the conversation so far?
3. Consistency: Is this turn's translation consistent with how earlier turns were translated (terminology, register, established facts)?

Format your output as a JSON object with a "results" array only (do not include completion_rate).
Each result should have: id (number), criteria (string), met (boolean), and reasoning (string).
\end{PromptBox}
\begin{PromptBox}{Multi-Turn Judge Prompt (Conversation)}
You are an expert linguistic and cultural evaluator, judging an ENTIRE interpreted two-party conversation for cross-turn (conversation-level) qualities: register/honorific consistency, cumulative meaning, and goal completion across the whole exchange.

Conversation Context: {conversation_context}

Full bilingual transcript:
{transcript_block}

[EDITORIAL -- {failed_turns_note} is empty unless a turn failed language verification; then: "Note: turn(s) [i, ...] failed language verification and should be weighed accordingly."]
{failed_turns_note}Verification Checklist (conversation-level -- judge the WHOLE conversation, not any single turn):
{verification_prompt}

For each item in the verification checklist, determine if the conversation as a whole successfully meets the criteria (Yes/No).
Also provide a brief reasoning for your decision, referencing specific turns where relevant.

Format your output as a JSON object with a "results" array only (do not include completion_rate).
Each result should have: id (number), criteria (string), met (boolean), and reasoning (string).
\end{PromptBox}

%% file: appendix/J_qualitative_examples.tex
\section{Qualitative Examples}
\label{sec:appendix-qualitative-examples}

Table~\ref{tab:mt-metric-blindspots} reports that 16.2--17.5\% of the Gemini 3.1 Pro interpreter's
translations a metric rates highest still fail our checklist. Table~\ref{tab:qualitative-blindspots}
shows four such cases from the 6,000-scenario correlation study (\S\ref{sec:analysis-mtmetrics}). In each row
CometKiwi places the output in its top quartile and both GEMBA-MQM and MQM-APE annotate zero errors, while
our checklist scores it 0.33--0.50. Each translation is fluent and semantically accurate, which the
reference-free metrics reward. What fails is a register, honorific, or face-management expectation of the
target language and relationship, which none of the three metrics scores. The rows span four mechanisms:
Korean speech level, Bengali pronoun choice, Indonesian formality, and Arabic register with grammatical
gender. All four are among the 113 outputs (1.9\% of 5,999) where CometKiwi's top quartile coincides with
zero errors from both MQM metrics and a checklist pass rate of at most 0.5.

\begin{table*}[t]
  \centering
  \caption{Four segments translated by the Gemini 3.1 Pro interpreter where CometKiwi ranks the output in its top
  quartile and GEMBA-MQM / MQM-APE annotate zero errors, yet our checklist
  passes only the fraction shown in \textbf{Chk.} (L1--L3 criteria met / total). \textbf{Kiwi} is the CometKiwi score. English glosses are in italics.
  Full checklists, per-criterion Clef probabilities, and Gemini 3.1 Pro reasoning are available in the supplementary data.}
  \label{tab:qualitative-blindspots}
  \small
  \newcommand{\qualimg}[1]{\raisebox{\dimexpr\ht\strutbox-\height\relax}{\includegraphics[width=\linewidth]{#1}}}
  \setlength{\tabcolsep}{4pt}
  \newcolumntype{L}[1]{>{\raggedright\arraybackslash}p{#1}}
  \begin{tabular}{@{}L{0.05\textwidth}L{0.215\textwidth}L{0.215\textwidth}c c L{0.33\textwidth}@{}}
    \toprule
    \textbf{Pair} & \textbf{Source} & \textbf{Translation} & \textbf{Kiwi} & \textbf{Chk.} & \textbf{What the metrics miss} \\
    \midrule

    ben \newline $\to$kor &
    \qualimg{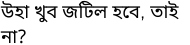} \newline
    \textit{``That would be very complicated, wouldn't it?''} &
    \qualimg{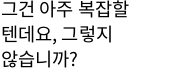} \newline
    \textit{``That would be very complicated, wouldn't it?'' (polite endings)} &
    0.89  &
    6/18 &
    A senior official asks a junior officer a rhetorical question. Uses upward-polite endings instead of the downward speech level a superior owes a
    subordinate in Korean, so the rhetorical question reads as a genuine one and loses the power dynamic and sarcastic \textit{nunchi} pressure.
    \\ \addlinespace[6pt]

    ind \newline $\to$ben &
    Kau bukan tahanan. Kau bebas bergabung dengan District 13. \newline
    \textit{``You're not a prisoner. You're free to join District 13.''} &
    \qualimg{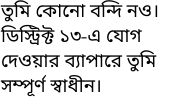} \newline
    \textit{``You (informal) are not a prisoner. You are completely free with regard to joining District 13.''} &
    0.91  &
    8/16 &
    A coordinator reassures a rescued, distrustful defector. Keeps informal where the vulnerable guest is owed respect, 
    and renders a warm invitation as an administrative fact, risking relational misexpectation. The listener's reply
    comes back confused.
    \\ \addlinespace[6pt]

    ben \newline $\to$ind &
    \qualimg{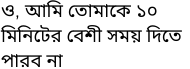} \newline
    \textit{``Oh, I can't give you (informal) more than 10 minutes.''} &
    Oh ya, saya tidak bisa memberikan kamu waktu lebih dari 10 menit. \newline
    \textit{``Oh right, I can't give you (informal) more than 10 minutes.''} &
    0.90  &
    6/16 &
    An executive speaks to a first-time student interviewer. Carries the informal register, and keeps the blunt refusal frame
    instead of a situational boundary. \\ \addlinespace[6pt]

    kor \newline $\to$arb &
    \qualimg{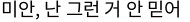} \newline
    \textit{``Sorry, I don't believe in that stuff.'' (informal)} &
    \qualimg{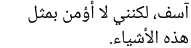} \newline
    \textit{``Sorry (masc.), but I do not believe in such things.'' (formal)} &
    0.90  &
    8/18 &
    A speaker turns down a close friend's suggestion. Renders informal Korean peer speech as formal Modern Standard Arabic,
    erasing the closeness between friends, and uses masculine agreement
    where the context (and the recipient's reply) establishes a female speaker. \\

    \bottomrule
  \end{tabular}
\end{table*}